\documentclass[12pt]{article}
\usepackage{latexsym}
\usepackage{times}
\usepackage{amsmath}
\usepackage{amssymb}
\usepackage{verbatim}
\usepackage{natbib}
\usepackage{multirow}
\usepackage{color}
\usepackage{float}

\usepackage[utf8]{inputenc} 
\usepackage[T1]{fontenc}    
\usepackage{hyperref}       
\usepackage{url}            
\usepackage{booktabs}       
\usepackage{microtype}      
\usepackage{bm}

\usepackage{algorithm}
\usepackage[noend]{algpseudocode}

\newif\ifpdf
\ifx\pdfoutput\undefined
  \pdffalse                     
  \usepackage[dvips]{graphicx}
\else
  \pdfoutput=1                  
  \pdftrue
  \usepackage[pdftex]{graphicx}
\fi

\def\x{{\mathbf{x}}}
\def\p{{\mathbf{p}}}

\def\z{{\mathbf{z}}}

\def\y{{\mathbf{y}}}
\def\t{{\mathbf{t}}}
\def\bfv{{\mathbf{v}}}
\def\bfy{{\mathbf{y}}}

\def\bpi{{\boldsymbol{\pi}}}

\def\bmu{{\boldsymbol{\mu}}}
\def\bfzero{{\boldsymbol{0}}}
\newcommand{\defeq}{{\stackrel{def}{=}}}

\def\bE{{\mathbb{E}}}
\def\bR{{\mathbb{R}}}

\newcommand{\cut}[1]{}

\begin{document}

\title{
\makebox[0pt][l]{\raisebox{1in}[0pt][0pt]{
\parbox{6in}{\small Final m/s version of paper accepted for
  publication in  \emph{Neural Computation}.}}}
Inference and Learning for Generative Capsule Models}

  \author{Alfredo Nazabal\thanks{Equal contribution.} \thanks{Work carried out while AN was at the Alan Turing Institute.} \\
       Amazon Development Centre Scotland \\
       Edinburgh, UK \\
       \texttt{alfrena@amazon.com}
       \and
       Nikolaos Tsagkas$^\ast$\thanks{Part of this work was
         carried out when NT was a MSc student at the University of
         Edinburgh.} \\
       School of Informatics,  University of Edinburgh, \\
             Edinburgh EH8 9AB, UK \\
       \texttt{n.tsagkas@ed.ac.uk}
       \and
       Christopher K. I. Williams \\
       School of Informatics,  University of Edinburgh, \\
             Edinburgh EH8 9AB, UK, and \\
             The Alan Turing Institute, London, UK \\
             \texttt{c.k.i.williams@ed.ac.uk} }

\maketitle

\begin{abstract}
Capsule networks (see e.g.~\citealp{hinton2018matrix}) aim to encode
knowledge of and reason about the relationship between an object and
its parts. In this paper we specify a \emph{generative} model for such
data, and derive a variational algorithm for inferring the
transformation of each model object in a scene, and the assignments of observed parts to
the objects. We derive a learning algorithm for
  the object models, based on variational expectation maximization
  \citep{jordan-ghahramani-jaakkola-saul-99}.
We also study an alternative inference algorithm
based on the RANSAC method of \citet{fischler1981random}.
We apply these inference methods to (i) data generated from multiple
geometric objects like squares and triangles (``constellations''), and
(ii) data from a parts-based model of faces.  Recent work
by~\cite{kosiorek2019stacked} has used amortized inference via stacked
capsule autoencoders (SCAEs) to tackle this problem---our results show
that we significantly outperform them where we can make comparisons
(on the constellations data).
\end{abstract}
  
\noindent Keywords: Capsules, variational inference, permutation matrix, Sinkhorn-Knopp
  algorithm, RANSAC.

\section{Introduction}
An attractive way to set up the problem of object recognition is
\emph{hierarchically}, where an object is described in terms of its
parts, and these parts are in turn composed of sub-parts, and so on.
For example a face can be described in terms of the eyes, nose, mouth,
hair, etc.; and a teapot can be described in terms of a body, handle,
spout and lid parts.  This approach has a long history in computer
vision, see e.g.\ the work on Pictoral Structures by
\citet{fischler-elschlager-73}, and
Recognition-by-Components by \citet{biederman-87}. More
recently \citet{felzenszwalb-girshick-mcallester-ramanan-09}
used discriminatively-trained parts-based models to obtain
state-of-the-art results (at the time) for object recognition.
Advantages of recognizing objects by first recognizing their constituent
parts include tolerance to the occlusion of some parts, and that parts
may vary less under a change of pose than the appearance of the whole
object.

Recent work by
\citet{sabour2017dynamic} and \citet{hinton2018matrix} 
has developed \emph{capsule networks}. These exploit the fact that if the
pose\footnote{i.e.\ the location and rotation of the object in 2D or
3D.} of the object changes, this can have very complicated effects on
the pixel intensities in an image, but the geometric transformation
between the object and the parts can be described by a simple linear
transformation (as used in computer graphics). In these papers
a part in a lower
level can vote for the pose of an object in the higher level, and an
object's presence is established by the agreement between votes for
its pose in a process called ``routing-by-agreement''.
\citet[p.~1]{hinton2018matrix}  describe this as a process which
``updates the probability with
which a part is assigned to a whole based on the proximity of the vote
coming from that part to the votes coming from other parts that are
assigned to that whole''.  Subsequently~\citet{kosiorek2019stacked}
framed inference for a capsule network in terms of an autoencoder, the
Stacked Capsule Autoencoder (SCAE).  Here, instead of the iterative
routing-by-agreement algorithm, a neural network $h^{\mathrm{caps}}$
takes as input the set of input parts and outputs predictions for the
object capsules' instantiation parameters $\{\y_k \}_{k=1}^K$.
Further networks $h_k^{\mathrm{part}}$ are then used to predict part
candidates from each $\y_k$.

The objective function used in~\citet{hinton2018matrix} (their eq.\ 4)
is quite complex (involving four separate terms), and is not derived
from first principles.  In this paper we argue that the description in
the paragraph above is backwards---we find it more natural to first
describe the
generative process \emph{by which an object gives rise to its parts},
and that the appropriate routing-by-agreement inference algorithm then
falls out naturally from this principled formulation.

The contributions of this paper are to:
\begin{itemize}
  
\item Derive a novel variational inference algorithm for
  routing-by-agreement, based on a generative model of object-part
  relationships, including a relaxation of the permutation-matrix
  formulation for matching object parts to observations;
\item Focus on the problem of inference for \emph{scenes containing
multiple objects}. Much of the work on capsules considers
  only single objects (although sec.\ 6 in \citet{sabour2017dynamic} and
 \citet[sec.\ 4.4]{ribeiro-leontidis-kollias-20b} are notable exceptions).
\item Demonstrate the effectiveness of our algorithm  on (i)
    ``constellations'' data generated from multiple geometric objects
    (e.g. triangles, squares) at arbitrary translations, rotations and
  scales;  and (ii) data of multiple faces from a novel parts-based
  model of faces;
  \item Evaluate the performance of our algorithm and the RANSAC
    method vs.\ competitors on the constellations data.
  \item Derive a learning algorithm for the object models, based on
    variational expectation maximization \citep{jordan-ghahramani-jaakkola-saul-99}.
\end{itemize}  

The structure of the remainder of the paper is as follows: in section
\ref{sec:overview} we provide an overview of the method.  Section
\ref{sec:generative} gives details of the generative model and the
matching distribution between observed and model parts.  The
variational inference algorithm derived from this model is given in
section \ref{sec:VI}, and RANSAC inference is described in
sec.\ \ref{sec:ransac}. Related work is discussed in section
\ref{sec:relwork}, and approaches to learning generative capsule
models are given in section \ref{sec:learning}.
Our experiments and results are described in section \ref{sec:expts_res}.
We conclude with a discussion.

\begin{figure}[t]
  \begin{center}
  \begin{tabular}{ccc}
    \includegraphics[width=0.5\linewidth]{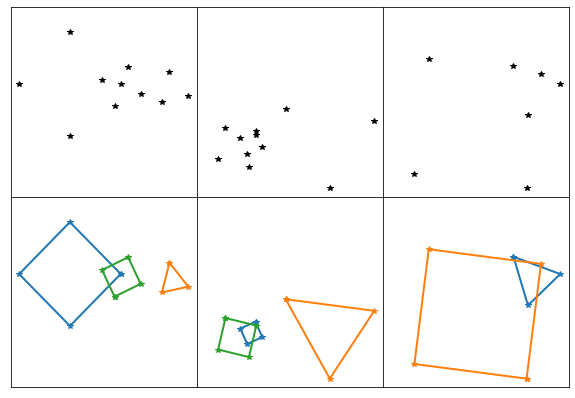} &
    \includegraphics[scale=0.25]{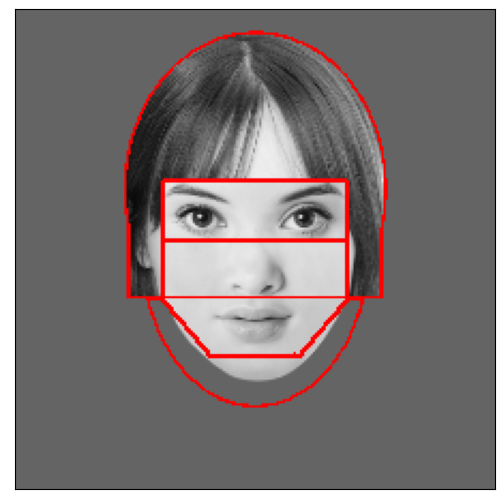} &
    \includegraphics[scale=0.25]{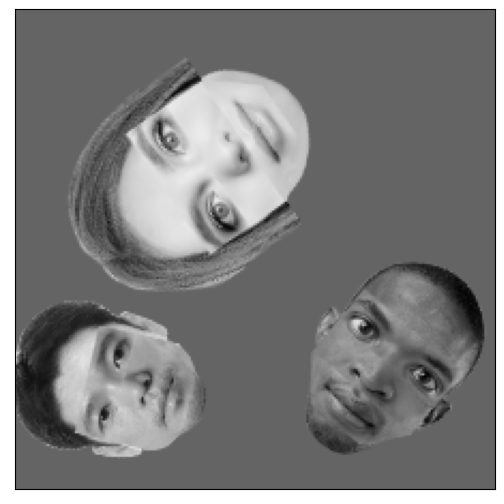}\\
    (a) & (b) & (c) 
  \end{tabular}
  \end{center}
    \caption{(a) Scenes composed of 2D points (upper
figures) and their corresponding objects (lower
figures). (b) A synthetic face. The red lines indicate the
  areas of the 5 part types (i.e. hair, eyes, nose, mouth and
  jaw). (c) Example scene with 3 randomly transformed faces.
    \label{fig:constandfaces}}
\end{figure}

\section{Overview} \label{sec:overview}
Consider images of a set of objects in different poses, such as
images of handwritten digits, faces, or geometric shapes in 2D
or 3D.  An object can be defined as an instantiation of a specific
model object (or template) along with a particular pose (or geometric
transformation).  Furthermore, objects, and thus templates, are
decomposed in parts, which are the basic elements that comprise the
objects.  For example,
faces can be decomposed into parts (e.g.\ mouth, nose etc.),
or a geometric shape can be decomposed into vertices. These parts
can have internal variability, (e.g.\ eyes open or shut).
  
More formally, let $T = \{T_k\}_{k=1}^K$ be the set of $K$ templates
that are used to generate a scene.  Each template
$T_k=\{\p_n\}_{n=1}^{N_k}$ is composed of $N_k$ parts $\p_n$.  We
assume that scenes can only be generated using the available
templates. Furthermore, every scene can present a different
configuration of objects, with some objects missing in some scenes.
For example, in scenes that could potentially contain all digits from
0 to 9 once, and if only the digits 2 and 5 are in the image, we consider
that the other digits are missing. If all the templates were
employed in the scene, then the number of observed parts $M$ is equal
to the sum of all the parts of all the templates $N = \sum_{k=1}^K
N_k$.

Each observed template $T_k$ in a scene is then transformed by an
independent transformation $\y_k$, different for each template,
generating a set of transformed parts $X_k = \{\x_n\}_{n=1}^{N_k}$
\begin{equation}
T_k \stackrel{\y_k}{\rightarrow} X_k .
\end{equation}
The transformation $\y_k$ includes both the geometric transformation of
the template, and also the appearance variability in the parts.

In this paper we assume that we are given a scene 
$X = \{\x_m\}_{m=1}^M$ composed of $M$ observed parts coming from multiple
templates. (For example, the Part Capsule Autoencoder (PCAE)
  of \citet{kosiorek2019stacked} outputs a set of parts.)
The \emph{inference problem} for $X$ involves a number of 
different tasks.  We need to determine which objects from the set of
templates were employed to generate the scene. 
Also, we need to infer what transformation
$\y_k$ was applied to each template to generate the objects.  This
allows us to infer the \emph{correspondences} between the
template parts and the scene parts.

We demonstrate our method on ``constellations'' data as shown in
Fig.\ \ref{fig:constandfaces}(a), and data containing multiple faces
(Fig.\ \ref{fig:constandfaces}(c)).  In the constellations data, the
real generators are triangle and square objects, but only their
vertices are provided in the data.  The faces data is generated from
the parts-based model of faces shown in
Fig.\ \ref{fig:constandfaces}(b) and described in
sec.\ \ref{sec:parts_face}.

\section{A Generative Capsules Model (GCM)}\label{sec:generative}
We propose a generative model to describe the problem.  Consider a
template (or model) $T_k$ for the $k$th object. $T_k$ is composed of
$N_k$ parts $\{ \p_n \}_{n=1}^{N_k}$.\footnote{For simplicity of
notation we suppress the dependence of $\p_n$ on $k$ for now.}  Each
part $\p_n$ is described in its reference frame by its \emph{geometry}
$\p^g_n$ and its \emph{appearance} $\p^a_n$.  Each object also has
associated latent variables $\y_k$ which transform from the reference
frame to the image frame, so $\y_k$ is split into geometric variables
$\y^g_k$ and appearance variables $\y^a_k$.

\paragraph{Geometric transformations:}
Here we consider 2D templates and a similarity transformation
(translation, rotation and scaling) for each object, but this can be
readily extended to allow 3D templates and a scene-to-viewer camera
transformation. We assume that $\p_n^g$ contains the $x$ and $y$
locations of the part, and also its size $s_n$ and orientation
$\phi_n$ relative to the reference frame.\footnote{For
the constellations data, the size and orientation information is not
present, nor are there any appearance features.}
The size and orientation are represented as the projected size of the
part onto the $x$ and $y$ axes, as this allows us to use linear
algebra to express the transformations (see below).
Thus $\p_n^g = (p^g_{nx}, p^g_{ny}, s_n \cos\phi_n, s_n \sin\phi_n)^T$.

Consider a template with parts $\p^g_n$ for $n = 1,
\ldots, N_k$ that we wish to scale by a factor $s$, rotate through
with a clockwise rotation angle $\theta$ and translate by $(t_x,t_y)$.
We obtain a transformed object with geometric observations for the
$n$th part $\x^g_n = (x^g_{nx},x^g_{ny},x^g_{nc}, x^g_{ns})$, where the $c$ and $s$
subscripts denote the projections of the scaled and rotated part onto
the $x$ and $y$ axes respectively ($c$ and $s$ are mnemonic for cosine and sine).

For each part in the template, the geometric transformation works as follows:
\begin{equation}\label{eq:nxycs}
\begin{pmatrix} x_{nx}\\ x_{ny} \\ x_{nc} \\ x_{ns}\end{pmatrix} =   
\begin{pmatrix} 1 & 0 & p_{nx} & p_{ny}\\
         0 & 1 & p_{ny} & -p_{nx} \\
         0 & 0 & s_n \cos\phi_n & - s_n \sin\phi_n \\
         0 & 0 & s_n \sin\phi_n & s_n \cos\phi_n \\    \end{pmatrix}
\begin{pmatrix}
t_x\\
t_y\\
s\cos{\theta}\\
s\sin{\theta}
\end{pmatrix}.
\end{equation}
Decoding the third equation, we see that $x_{nc} = s_n s
\cos\phi_n \cos \theta - s_n s \sin\phi_n \sin \theta =
s_n s \cos(\phi_n + \theta)$ using standard trigonometric identities.
The $x_{ns}$ equation is derived similarly. We shorten eq.\ \ref{eq:nxycs} to
$\x^g_n = F^g_{kn} \y^g_k$,
where $\y^g_k$ is the $\bR^4$ column vector, and $F^g_{kn} \in \bR^{4 \times 4}$
is the matrix to its left.\footnote{Here $F$ is mnemonic for ``features''.}
Allowing Gaussian observation noise with
precision $\lambda$ we obtain
\begin{equation}
p(\x^g_n|T_k, \y^g_k) \sim
N\left(\x^g_{n}|F^g_{kn}\y^g_{k},\lambda^{-1}I\right).
\label{eq:pxg_supp}
\end{equation}

The prior distribution over similarity transformations $\y^g_k$ is
modelled with a $\bR^4$ Gaussian distribution with mean $\bmu^g_0$ and
covariance matrix $D^g_0$: 
\begin{equation}
\label{eq:p_Y}
p(Y^g) = \prod_{k=1}^K N(\y^g_{k}|\bmu^g_0,D^g_0),
\end{equation}
where $Y^g$ denotes the set $\{ \y^g_k \}_{k=1}^K$.
Notice that modelling $\y^g_k$ with a Gaussian distribution implies that
we are modelling the translation $(t_x,t_y)$ in $\bR^2$ with a
Gaussian distribution.  If $\bmu^g_0=\mathbf{0}$ and $D^g_0 = I_4$ 
then $s^2 = (y^g_{k3})^2 + (y^g_{k4})^2$
has a $\chi^2_2$ distribution, and
$\theta = \arctan{{y^g_{k4}}/{y^g_{k3}}}$ is uniformly distributed
in its range $[-\pi,\pi]$ by
symmetry.  For more complex linear transformations (e.g.\ an affine
transformation), we need only to increase the dimension of $\y^g_k$ and
change the form of $F^g_{kn}$, but the generative model
in~\eqref{eq:p_Y} would remain the same. For the 3D case, note that
  \citet{basri-96} has shown that 
every affine projection of an object represents some uncalibrated
paraperspective projection of the object.

\paragraph{Appearance transformations:}
The appearance $\x_n^a$ of part $n$ in the image depends on 
$\y^a_k$. For our faces data, $\y^a_k$ is a vector latent variable
which models the co-variation of the appearance of the parts via
a linear (factor analysis) model; see sec.\ \ref{sec:parts_face} for a fuller
description.
Hence
\begin{equation}
p(\x^a_n|T_k, \y_k) \sim
N\left(\x^a_{n}|F^a_{kn} \y^a_{k} + \boldsymbol{m}_{kn}^a, D^a_{kn}\right),
\label{eq:pxa_supp} 
\end{equation}
where $F^a_{kn}$ maps from $\y^a_k$ to the predicted appearance features
in the image, $D^a_{kn}$ is a diagonal matrix of variances and
$\boldsymbol{m}_{kn}^a$ allows for the appearance features to have a
non-zero mean. The dimensionality of the $n$th part of the $k$th
template is $d_{kn}$. The prior for $\y^a_k$ is taken to be a standard
Gaussian, i.e.\ $N(\mathbf{0},I)$. Combining (\ref{eq:p_Y}) and the
prior for $\y^a_k$, we have that $p(\y_k)  = N(\bmu_0, D_0)$, where
$\bmu_0$ stacks $\bmu^g_0$ and $\mathbf{0}$ from the appearance,
and $D_0$ is a diagonal matrix with blocks $D^g_0$ and $I$.

\paragraph{Joint distribution:}
Let $z_{mnk} \in \{0,1\}$ indicate whether observed part $\x_m$
matches to part $n$ of object $k$. The set of these match variables is
denoted by $Z$.  Every observation $m$ belongs uniquely to a tuple
$(k,n)$, or in other words, a point $\x_m$ belongs uniquely to the
part defined by $\y_k$ acting on the template matrix $F_{kn}$. The
opposite is also partially true; every tuple $(k,n)$ belongs uniquely
to a point $m$, or it is unassigned if part $n$ of template $k$ is
missing in the scene.

The joint distribution of the variables in the model is given by
\begin{equation}
p(X,Y,Z) = p(X|Y,Z)p(Y)p(Z),
\end{equation}
where $p(X|Y,Z)$ is a Gaussian mixture model explaining how the points in a scene were generated from the templates
\begin{equation}\label{eq:p_X}
p(X|Y,Z) = \prod_{m=1}^M\prod_{k=1}^K\prod_{n=1}^{N_k}
N\left(\x_{m}|F_{kn}\y_{k} + \boldsymbol{m}_{kn}, D_{kn} \right)^{z_{mnk}},
\end{equation}
where $D_{kn}$ consists of the diagonal matrices
$\lambda^{-1} I$ and $D^a_{kn}$ and $\boldsymbol{m}_{kn}$
consists of a zero vector for the geometric features stacked
on top of the mean for the appearance features
$\boldsymbol{m}^a_{kn}$.
Note that $F_{kn}$ has blocks of
zeros so that $\x^g_m$ does not depend on $\y^a_{k}$,
and $\x^a_m$ does not depend on $\y^g_{k}$.

\paragraph{Annealing parameter:} During inference, where the
model is fitted to  data, it is useful to modify the covariance matrix $D_{kn}$ to
$\beta^{-1} D_{kn}$, where $\beta$ is a parameter $< 1$. The effect of
this is to inflate the variances in $D_{kn}$, allowing greater
uncertainty in the inferred matches early on in the fitting process,
as used, e.g.\ in \citet{revow-williams-hinton-96}.  $\beta$ is
increased according to an annealing schedule during the fitting.

\paragraph{Match distribution $p(Z)$:}
In a standard Gaussian mixture model, the assignment matrix $Z$ is
characterized by a Categorical distribution, where each point $\x_m$
is assigned to one part
\begin{equation}
\label{eq:p_Z}
p(Z) = \prod_{m=1}^M \text{Cat}(\z_{m}|\bpi),
\end{equation}
with $\z_m$ {being a 0/1 vector with only one 1,}
and $\bpi$ being the probability vector for each tuple $(k,n)$.
However, the optimal solution to our problem occurs when each part of
a template belongs uniquely to one observed part in a scene.  This means that
$Z$ should be a permutation matrix, where each point $m$ is assigned
to a tuple $(k,n)$ and vice versa.
Notice that a permutation matrix is a square matrix, so if $M \leq N$,
we add dummy rows to $Z$, which are assigned to missing points in the
scene.

The set of permutation matrices of dimension $N$ is a discrete set
containing $N!$ permutation matrices. 
A discrete prior over permutation matrices assigns each valid matrix
  $Z_i$ a probability $\pi_i$:
\begin{equation}
\label{eq:p_Z_true}
p_{perm}(Z) = \sum_{i=1}^{N!}  \pi_i \; I[Z=Z_i] 
\end{equation}
with $\sum_{i=1}^{N!} \pi_i = 1$ and $I[Z=Z_i]$ being the indicator
function, equal to $1$ if $Z=Z_i$ and $0$ otherwise.

The number of possible permutation matrices increases
as $N!$, {which makes exact inference over permutations intractable,
except for very small $N$.}
An interesting property of $p_{perm}(Z)$ is that its first moment
$\bE_{p_{perm}}[Z]$ is a doubly-stochastic (DS) matrix, a matrix of
elements in $[0,1]$ whose rows and columns sum to 1.
We propose to relax $p_{perm}(Z)$ to a distribution $p_{DS}(Z)$ that
is characterized by the doubly-stochastic matrix $A$ with elements
$a_{mnk}$, such that $\bE_{p_{DS}}[Z]=A$:
\begin{equation}
\label{eq:p_Z_approx}
p_{DS}(Z) =  \prod_{m=1}^{N}\prod_{k=1}^{K}\prod_{n=1}^{N_k} a_{mnk}^{z_{mnk}}.
\end{equation}
$A$ is fully characterized by $(N-1)^2$ elements. In the absence of
any prior knowledge of the affinities, 
a uniform prior over $Z$ with $a_{mnk}=\frac{1}{N}$ can be used.
However, note that $p_{DS}$ can
also represent a particular permutation matrix $Z_i$ by setting the
appropriate entries of $A$ to 0 or  1, and indeed we would expect this to occur during
variational inference (see sec.\ \ref{sec:VI}) when the model
converges to a correct solution.

\paragraph{Related models:} Our model for a single object has both
geometric and appearance variability (see eqs.\ \ref{eq:pxg_supp} and
\ref{eq:pxa_supp}). A similar model but with geometric features only
was developed by~\citet{revow-williams-hinton-96}.
\citet{fergus-perona-zisserman-03} described a ``constellation of
parts'' model, that used a joint Gaussian model for locations of the
parts, and an image-patch model for each part appearance. However, the
appearance model was a single Gaussian per part, without the
correlations between parts afforded by the factor analysis model.
This model was applied to images of single (foreground) objects,
summing out over possible assignments $Z$.
\citet{rao-ballard-99} developed 
a hierarchical factor analysis model, but used it to model
extended edges in natural image patches rather than correlations
between the parts of an object.
See sec.\ \ref{sec:learning} for further discussion of
parts-based models.

\paragraph{Hierarchical modelling:} above we have described
a two-layer model with part-object relations. This can of course be
extended to a deeper hierarchy; for example we would expect to find
relationships between the objects in a scene, such as the
relationships between a dining table and the dining chairs, or between
a bed and its associated nightstand(s). Such scene-level relationships
can be formulated in terms of graphical models for the groups of
objects. We believe the most difficult aspect is handling the
assignment $Z$ between model parts and observed parts (as covered in
this paper), and that inference in the graphical model above is fairly
standard, and is left for future work.

\section{Variational Inference}\label{sec:VI}

Variational inference for the above model can be derived similarly to
the Gaussian Mixture model case~\citep[Sec.\ 10.1]{bishop2006pattern}.
The variational distribution under the mean
field assumption is given by $q(Z,Y) = q(Z)q(Y)$,
The evidence lower error bound (ELBO) for this model is derived in
  Appendix \ref{sec:appVI}. Optimizing the ELBO with respect to either
$q(Z)$ or $q(Y)$, the  optimal solutions  can be expressed as
\begin{eqnarray}
\log{q(Z)} &\propto & \bE_{q(Y)}[\log{p(X,Y,Z)}] ,\\
\log{q(Y)} &\propto & \bE_{q(Z)}[\log{p(X,Y,Z)}] .
\end{eqnarray}
These alternating updates of $q(Y)$ and $q(Z)$ carry out routing-by-agreement.

For $q(Z)$ we obtain an expression with the same form
as the prior in~\eqref{eq:p_Z_approx}
\begin{eqnarray}
q(Z) \propto \prod_{m=1}^N\prod_{k=1}^K\prod_{n=1}^{N_k} \rho_{mnk}^{z_{mnk}},\label{eq:q_Z}
\end{eqnarray}
where $\rho_{mnk}$ represents the unnormalized probability of point
$m$ being assigned to tuple $(k,n)$ and vice versa.  These
unnormalized probabilities have a different form depending on whether
we are considering a point that appears in the scene ($m \leq M$)
\begin{align}
&\log{\rho_{mnk}} = \log{a_{mnk}}-\frac{1}{2}\log{|\beta^{-1} D_{kn}|}- \frac{d_{kn}}{
2}\log{2\pi} - \nonumber\\
&\frac{\beta}{2}\bE_{\y_{k}}[(\x_{m}-F_{kn}\y_{k}-\boldsymbol{m}_{kn})^T D_{kn}^{-1} (
\x_{m}-F_{kn}\y_{k}-\boldsymbol{m}_{kn})],\label{eq:log_rho}
\end{align}
or whether we are considering a dummy row of the prior ($m > M$),
\begin{align}
\log{\rho_{mnk}} &= \log{a_{mnk}}. \label{eq:log_rho_dummy}
\end{align}
When a point is part of the scene~\eqref{eq:log_rho}, and thus $m \leq
M$ the update of $\rho_{mnk}$ is similar to the Gaussian mixture model
case.  However, if a point is not part of the
scene~\eqref{eq:log_rho_dummy}, and thus $m > M$ then the matrix is
not updated and the returned value is the prior $a_{mnk}$.
The expectation term in~\eqref{eq:log_rho} is given by:
\begin{multline}
\bE_{\y_{k}}[(\x_{m}-F_{kn}\y_{k}-\boldsymbol{m}_{kn})^T D_{kn}^{-1}
  (\x_{m}-F_{kn}\y_{k}-\boldsymbol{m}_{kn})]  = \\
(\x_{m} - F_{kn}\bmu_{k}- \boldsymbol{m}_{kn})^T D_{kn}^{-1}(\x_{m} -
F_{kn}\bmu_{k} - \boldsymbol{m}_{kn})  
+ \text{trace}(F_{kn}^TD_{kn}^{-1}F_{kn}\Lambda_{k}^{-1}).
\end{multline}

The normalized distribution $q(Z)$ becomes:
\begin{eqnarray}
q(Z) = \prod_{m=1}^N\prod_{k=1}^K\prod_{n=1}^{N_k} r_{mnk}^{z_{mnk}},\label{eq:R_doubleStochastic}
\end{eqnarray}
where $\bE_{q(Z)}[z_{mnk}] = r_{mnk}$.
The elements $r_{mnk}$ of matrix $R$ represent the posterior probability of each point $m$ being uniquely assigned to the part-object tuple $(n,k)$ and vice-versa.
This means that $R$ needs to be a DS matrix.  This can be achieved by
employing the Sinkhorn-Knopp algorithm~\citep{sinkhorn1967concerning}
as given in Algorithm \ref{alg:sinkhorn_knopp}, which updates a
square non-negative matrix by normalizing the rows and columns
alternately until the resulting matrix becomes doubly stochastic.

The use of the Sinkhorn-Knopp algorithm for approximating matching
problems has also been described by \cite{powell2019computing} and 
\cite{mena2020sinkhorn}, but note that in our case we also need to
alternate with inference for $q(Y)$.

\begin{algorithm}[h]
\caption{Sinkhorn-Knopp algorithm, taking as input a square non-negative matrix $M$}
\label{alg:sinkhorn_knopp}
\begin{algorithmic}[1]
\Procedure{SinkhornKnopp}{$M$} 
        \While{$M$ not doubly stochastic}
                \State Normalize rows of $M$: $m_{ij} = \frac{m_{ij}}{\sum_{j} m_{ij}}$, $\forall{i}$
                \State Normalize columns of $M$: $m_{ij} = \frac{m_{ij}}{\sum_{i} m_{ij}}$, $\forall{j}$
        \EndWhile
        \Return $M$
\EndProcedure
\end{algorithmic}
\end{algorithm}

Furthermore, the optimal solution to the assignment problem occurs
when $R$ is a permutation matrix itself. When this happens we
exactly recover a discrete posterior (with the same form
as~\eqref{eq:p_Z_true}) over permutation matrices where one of them
has probability one, with the others being zero.

The distribution for $q(Y)$ is a Gaussian with
\begin{align}
q(Y) &= \prod_{k=1}^K N(\y_k|\bmu_k,\Lambda_k^{-1}), \\\label{eq:Lambda_k}
\Lambda_{k} &= D^{-1}_{0} +\beta \sum^M_m \sum^{N_k}_n r_{mnk}F_{kn}^{T}D_{kn}^{-1} F_{kn}, \\\label{eq:mu_k}
\bmu_{k} &= \Lambda_k^{-1} \left[D_{0}\bmu_{0} + \beta\sum^M_m\sum^{N_k}_nr_{mnk}F_{kn}^{T}D_{kn}^{-1}(\x_{m}-\boldsymbol{m}_{kn})\right],
\end{align}
where the updates for both $\Lambda_k$ and $\bmu_{k}$ depend explicitly
on the annealing parameter $\beta$ and the templates employed in the
model.
Note that the prediction from datapoint $m$ to the mean of
$\y_k$ depends on
$r_{mnk}F_{kn}^{T} D^{-1}_{kn}(\x_{m}-\boldsymbol{m}_{kn})$, i.e.\ a
weighted sum of the predictions of each part $n$ with weights
$r_{mnk}$.
These expressions remain the same when considering a Gaussian mixture
prior such as~\eqref{eq:p_Z}.

Algorithm~\ref{alg:variational_inference} summarizes the
inference procedure for this model.

\begin{algorithm}[t]
\caption{Variational Inference}
\label{alg:variational_inference}
\begin{algorithmic}[1]
\State Initialize $\beta$, $\beta_{max}$ and $R \sim U[0,1]^{N\times N}$, $\forall m,n,k$
\State $R = \text{SinkhornKnopp}(R)$
\While{not converged}
        \State Update $\Lambda_k$~\eqref{eq:Lambda_k}, $\forall k$
        \State Update $\bmu_k$~\eqref{eq:mu_k}, $\forall k$
        \State Update $\log{\rho_{mnk}}$~\eqref{eq:log_rho}\eqref{eq:log_rho_dummy}, $\forall m,n,k$
        \State Update $R = \text{SinkhornKnopp}(\rho)$
        \If{ELBO has converged}
                \If{$\beta < \beta_{max}$}
                        \State Anneal $\beta$
                \Else
                        \State converged = True
                \EndIf
        \EndIf
\EndWhile
\Return $R, \{ \bmu_{k},\Lambda_k \}$
\end{algorithmic}
\end{algorithm}

\subsection{Comparison with other objective functions} \label{sec:VIvomparsion}
In \cite{hinton2018matrix} an objective function $cost^h_k$ is defined
(their eq.\ 1) which considers inference for the pose of a
higher-level capsule $k$ on pose dimension $h$.  Translating to our
notation, $cost^h_k$ combines the predictions $\bfv_{mk}$ from
each datapoint $m$ for capsule $k$ as $cost^h_k = \sum_m r_{mk} \ln
P^h_{m|k}$, where $P^h_{m|k}$ is a Gaussian, and $r_{mk}$ is the
``routing softmax assignment'' between $m$ and $k$.  It is interesting
to compare this with our equation \eqref{eq:mu_k}.  Firstly, note that
the vote of $\x_m$ to part $n$ in object $k$ is given explicitly by
$\beta \Lambda_k^{-1} F^T_{kn} D^{-1}_{kn} (\x_m - \boldsymbol{m}_{kn}),$ i.e.\ we do not
require the introduction of an independent voting mechanism, this
falls out directly from the inference.  Secondly, note that our $R$
must keep track not only of assignments to objects, but also to
\emph{parts} of the objects. In our experiments with the
  constellations data each observed part could match any object/part
  combination of the models, so this is necessary. For the faces data,
  the observed parts are of identifiable type (e.g.\ nose, mouth), so
  in this case they only need to vote for the object.  In contrast to
\citet{hinton2018matrix}, our inference scheme is derived from
variational inference on the generative model for $X$, rather than
introducing an \emph{ad hoc} objective function that corresponds
  to a clustering in $\bfy$-space.

\citet{ribeiro-leontidis-kollias-20} develop a variational Bayes
extension of the model of \cite{hinton2018matrix}. They write
down a mixture model in $\bfy$-space where the ``datapoints'' are
the votes $\bfv_{mk}$, and use Bayesian priors for the
mixing proportions, and for the means and covariance matrices of
the components. This can improve training stability, e.g.\ by
reducing the problem of variance-collapse where a capsule claims
sole custody of of a datapoint.
However, this is still a model for clustering in $\bfy$-space,
and not a generative model for $X$.

The specialization of the SCAE method of \citet{kosiorek2019stacked}
  to constellation data is called the ``constellation capsule
  autoencoder (CCAE)'' and discussed in their sec.\ 2.1.
Under their equation 5, we have that
\begin{align}\label{eq:CCAE_loss}
p(\x_{1:M}) = \prod_{m=1}^M\sum_{k=1}^K\sum_{n=1}^N \frac{a_ka_{k,n}}{\sum_i a_i \sum_j a_{ij}} N(\x_m|\mu_{k,n},\lambda_{k,n}),
\end{align}
where $a_k \in [0,1]$ is the presence probability of capsule $k$,
$a_{k,n} \in [0,1]$ is the conditional probability that a given
candidate part $n$ exists, and $\mu_{k,n}$ is the predicted location
of part $k,n$.
The $a_k$s are predicted by the network $h^{\mathrm{caps}}$, while the
$a_{k,n}$s and $\mu_{k,n}$s are produced by separate networks
$h_k^{\mathrm{part}}$ for each part $k$.

We note that \eqref{eq:CCAE_loss} provides an autoencoder style
reconstructive likelihood for $\x_{1:M}$, as the $a$'s and $\mu$'s
depend on the data.  To handle the arbitrary number of datapoints $M$,
the network $h^{\mathrm{caps}}$ employs a Set Transformer
architecture~\citep{lee-lee-kosiorek-choi-teh-19}.  In comparison to
our iterative variational inference, the CCAE is a ``one shot''
inference mechanism.  This may be seen as an advantage, but in scenes
with overlapping objects,
humans may perform reasoning like ``if that point is one of the
vertices of a square, then this other point needs to be explained by a
different object'' etc, and it may be rather optimistic to believe
this can be done in a simple forward pass.  Also, the CCAE cannot exploit
prior knowledge of the geometry of the objects, as it relies on an
opaque network $h^{\mathrm{caps}}$ which requires extensive training.

\begin{algorithm}[t]
\caption{RANSAC approach \label{alg:geometric_hashing}}
\begin{algorithmic}[1]
\State $T$: $K$ templates of the scene
\State $B_k$: base matrix for template $T_k$
\State $X$: $M$ points of the scene
\State $out = []$
        \For{$\x_i \in X$}
                \For{$\x_j \in X \setminus x_i$}
                        \State $\x_{ij} = \text{Vectorize}(\x_i,\x_j)$
                        \For{$k = 1:K$}
                                \State $\hat{\y}_k = B_k^{-1}\x_{ij}$
                                \State $T_k \xrightarrow{\hat{\y}_k} \hat{X}_k$
                                \If{ \text{SubsetMatch}($\hat{X}_k, X)$}
                                        \State Add $(T_k,\hat{\y}_k,\hat{X}_k)$ to $out$
                                \EndIf
                        \EndFor
                \EndFor
        \EndFor
\Return $out$
\end{algorithmic}
\end{algorithm}

\section{A RANSAC Approach to Inference} \label{sec:ransac}
A radical alternative to ``routing by agreement'' inference is to make
use of a ``random sample consensus'' approach (RANSAC,
\citealp*{fischler1981random}), where a minimal number of parts are
used in order to instantiate \cut{the whole} an object. The
original RANSAC fitted just one object, but Sequential RANSAC (see,
  e.g., \citealt*{torr-98,vincent-laganiere-01}) repeatedly removes
  the parts associated with a detected object and re-runs RANSAC, so
  as to detect all objects.

For the constellations problem, we can try matching any pair of points
on one of the templates to every possible pair of $M(M-1)$ points in
the scene.  The key insight is that a pair of known points is
sufficient to estimate the 4-dimensional $\hat{\y}_k$ vector in the
case of similarity transformations.  Using the transformation
$\hat{\y}_k$, we can then \emph{predict} the location of the remaining
parts of the template, and \emph{check} if these are actually present.
If so, this provides evidence for the existence of $T_k$ and
$\hat{\y}_k$.  After considering the $M(M-1)$ subsets, the algorithm
then combines the identified instantiations to give an overall
explanation of the scene.

More details of the RANSAC approach are given in Algorithm
\ref{alg:geometric_hashing}. Assume we have chosen parts $n_1$ and
$n_2$ as the basis for object $k$, and that we have selected
datapoints $\x_i$ and $\x_j$ as their hypothesized counterparts in the
image.  Let $\x_{ij}$ be the vector obtained by stacking $\x_i$ and
$\x_j$, and $B_k$ be the $4\times 4$ square matrix obtained by
stacking $F_{k n_1}$ and $F_{k n_2}$.  Then $\hat{\y}_k =
B_{k}^{-1}\x_{ij}$.  Finally, $\text{SubsetMatch}(\hat{X}_k, X)$
selects those points in $X_k$ that are close to $X$ with a given
tolerance and add them to the output. Among them, the solution is
given by the one that minimizes $\sum_{n=1}^{N_k}(\hat{x}_{nk} -
x_{nk})^2$.

The above algorithm chooses a specific basis for each object, but one
can consider all possible bases for each object.  It is then efficient
to use a hash table to store the predictions for each part, as used
in Geometric Hashing~\citep{wolfson1997geometric}.  Geometric Hashing
is traditionally employed in computer vision to match geometric
features against previously defined models of such features.  This
technique works well with partially occluded objects, and is
computationally efficient if the basis dimension is low.

For the faces data, each part has location, scale and orientation
information, so a single part is sufficient to instantiate the whole
object geometrically. For the appearance, we follow the procedure
for inference in a factor analysis model with missing data, as
given in \citet{williams-nash-nazabal-19}, to predict $\y^a_k$ given
the appearance of the single part.

It is interesting to compare the RANSAC and routing-by-agreement
approaches to identifying objects. In RANSAC a minimal basis is chosen
so as to instantiate the object, which is then verified by
finding the remaining parts in the predicted locations.
In contrast routing-by-agreement makes predictions from all
of the observed parts, and then seeks to adjust the $R$ matrix so as to
cluster these predictions into coherent objects. This is
generally an \emph{iterative process}, although the SCAE
autoencoder architecture is ``one shot''.\footnote{One can extend
variational autoencoders to use iterative amortized inference, see
\citet{marino-yue-mandt-18}.}

\section{Related Work} \label{sec:relwork}
The origins of capsule networks (including the name ``capsule'')
can be traced back at least as far as
the work of \citet{hinton-krizhevsky-wang-11} on Transforming
Auto-encoders. Capsules were further developed in 
\citet{sabour2017dynamic}. Above
we have already discussed later work by Hinton and co-workers, including
\cite{hinton2018matrix}, \citet{kosiorek2019stacked}, and also
\citet{ribeiro-leontidis-kollias-20}.

The paper by \citet{ribeiro-duarte-everett-leontidis-shah-22}
provides a thorough survey of capsules. Below we highlight
a few papers on capsules; for example \citet{li-zhu-19}
develop a capsule Restricted Boltzmann Machine (RBM). This generative
model contains a number of capsules, each having a binary existence
variable and a vector of capsule variables similar to our
$\bfy_k$. However, these capsule variables are not used to model
geometric transformations (the key idea behind capsule networks), but
instead to handle appearance variability of the parts.

Both \citet{li-zhu-naud-xi-20} and
\citet{smith-schut-gal-vanderwilk-21} propose a layered directed
generative model. These both make use of the "single parent
constraint", where a capsule in a layer can be connected to only one
capsule (its parent) in the layer above. Such tree-structured
relationships are similar to those studied in
\citet{hinton-ghahramani-teh-00} and \citet{storkey-williams-03}.
Like \citet{li-zhu-19}, \citet{li-zhu-naud-xi-20} do not explicitly
consider modelling geometric transformations with their network.
This aspect is explicit in \citet{smith-schut-gal-vanderwilk-21},
where their capsule variables (similar to our $\bfy_k$s) do model
pose, but not appearance variability.

\begin{figure*}[t]
    \centering
    \begingroup
    \setlength{\tabcolsep}{1pt} 
    \renewcommand{\arraystretch}{1} 
    
    \begin{tabular}{lcccccccc}
    (a) &
    \includegraphics[scale=0.165]{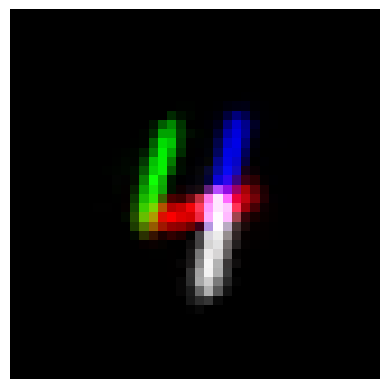} &
    \includegraphics[scale=0.165]{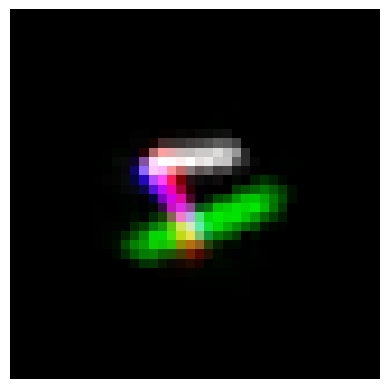} & 
    \includegraphics[scale=0.165]{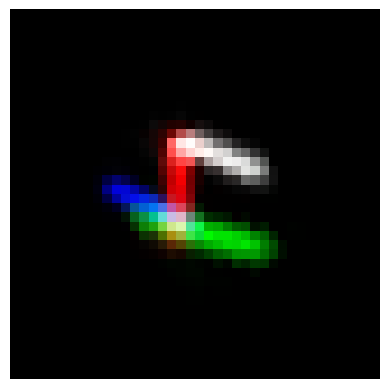} &
    \includegraphics[scale=0.165]{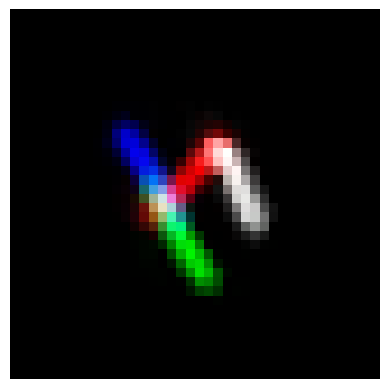} &
    \includegraphics[scale=0.165]{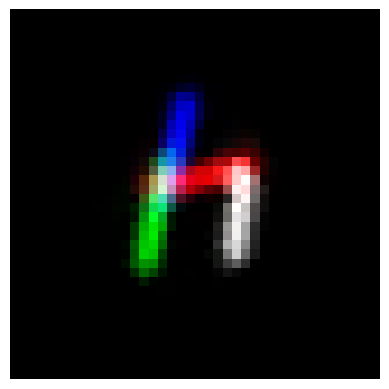} &
    \includegraphics[scale=0.165]{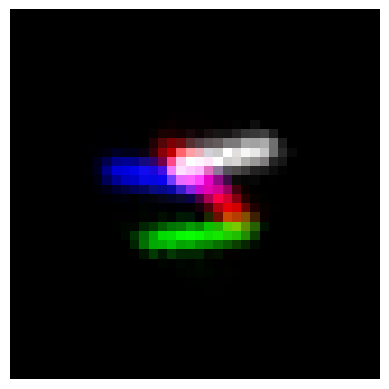} &
    \includegraphics[scale=0.165]{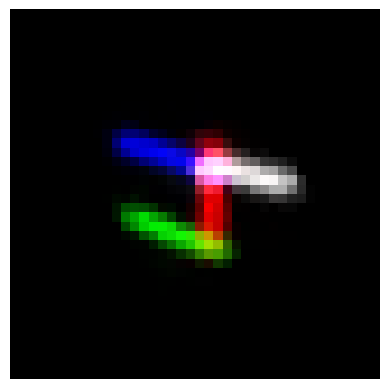} &
    \includegraphics[scale=0.165]{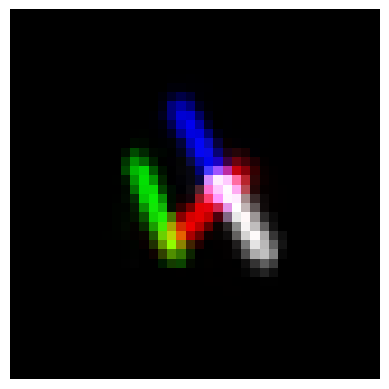} \\
    
    (b) &
    \includegraphics[scale=0.165]{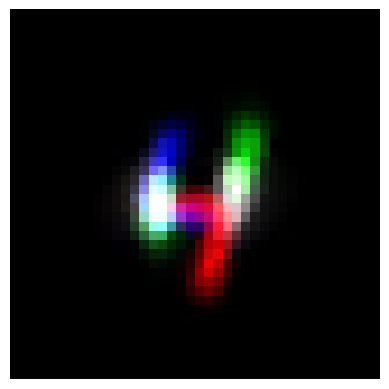} &
    \includegraphics[scale=0.165]{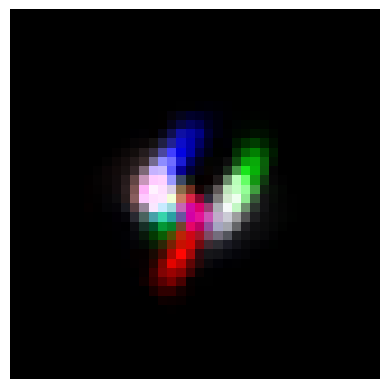} & 
    \includegraphics[scale=0.165]{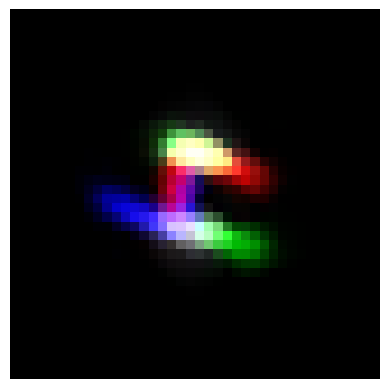} &
    \includegraphics[scale=0.165]{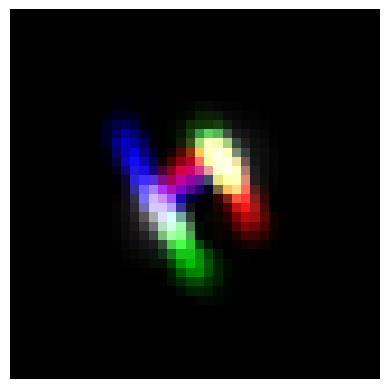} &
    \includegraphics[scale=0.165]{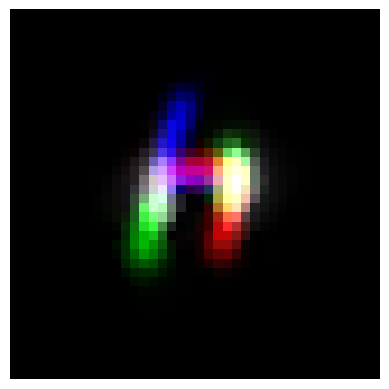} &
    \includegraphics[scale=0.165]{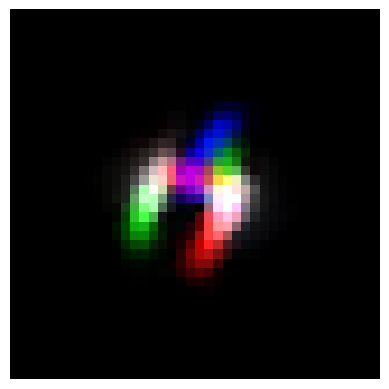} &
    \includegraphics[scale=0.165]{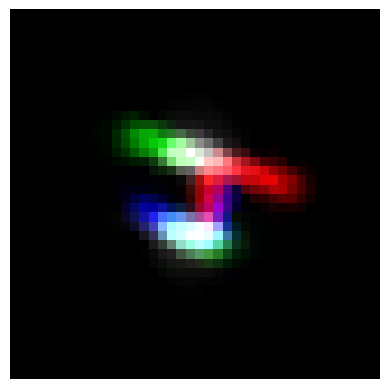} &
    \includegraphics[scale=0.165]{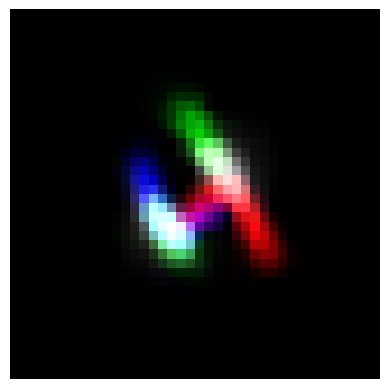} \\
    
    & $0^\circ$ &$45^\circ$ &$90^\circ$ &$135^\circ$ &$180^\circ$
    &$225^\circ$ &$270^\circ$ &$315^\circ$ \\ \\
    \end{tabular}
    \begin{tabular}{lcccccccc}
   \includegraphics[scale=0.16]{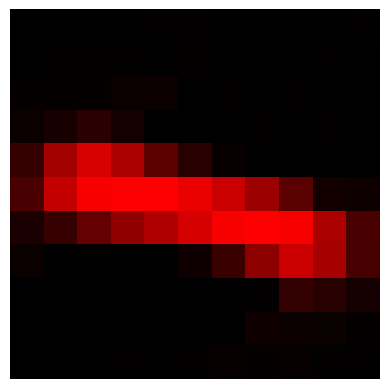} & 
   \includegraphics[scale=0.16]{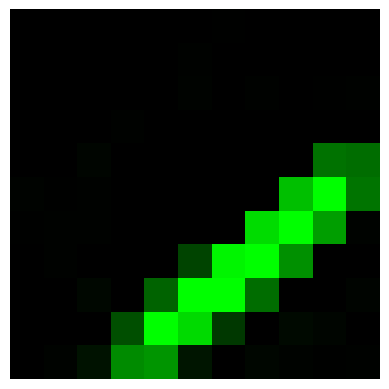} &
   \includegraphics[scale=0.16]{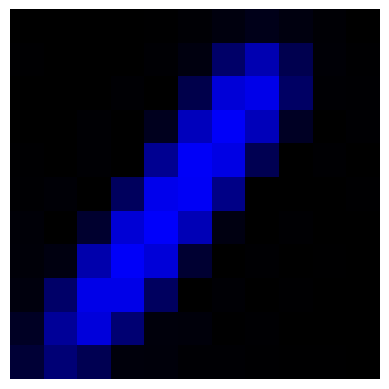} &
   \includegraphics[scale=0.16]{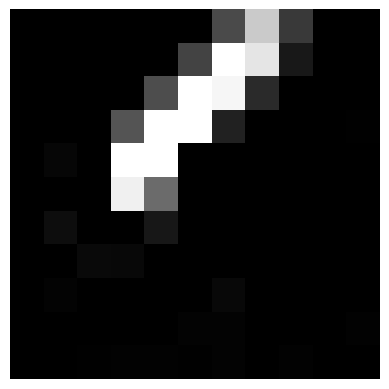}
   & \hspace*{10mm}& 
  \includegraphics[scale=0.16]{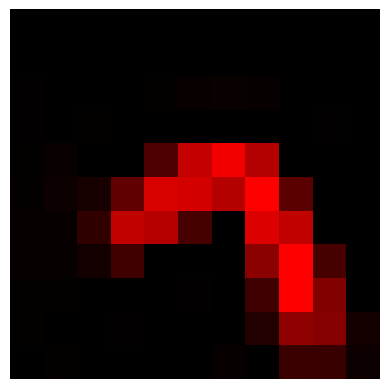} &
  \includegraphics[scale=0.16]{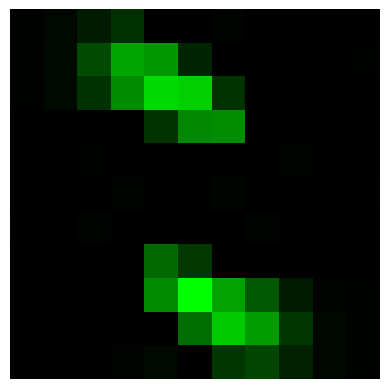} &  
  \includegraphics[scale=0.16]{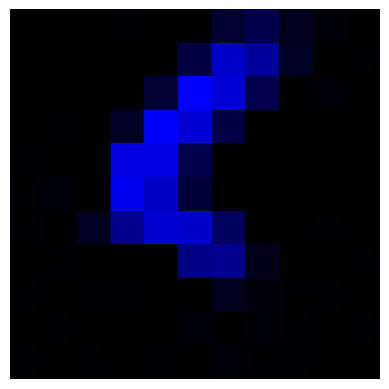} &  
  \includegraphics[scale=0.16]{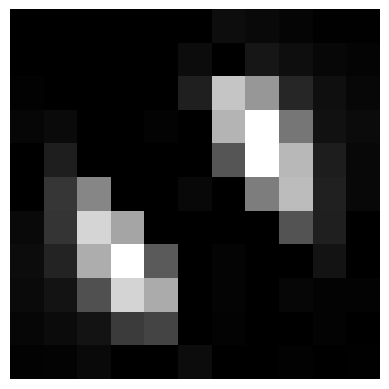} \\
  & & {\small part-set-a} & & & & & {\small part-set-b} &  
    \end{tabular}
    \endgroup
    \caption{PCAE-based reconstruction for different angles of
      rotation of the input scenes. Row (a) corresponds to
      reconstructions with the learned part-set-a and row (b)
      reconstructions with the learned part-set-b. The parts
    are $11 \times 11$ pixels, and the images in rows (a) and (b) are $40
    \times 40$.}
    \label{fig:supp_PCAE_equiv}
\end{figure*}

However, most importantly, these recent capsule models do not properly
handle the fact that the input to a capsule should be a \emph{set} of
parts; instead in their work the first layer capsules that interact
with the input image model specific location-dependent
features/templates in the image, and their second layer capsules have
interactions with the specific first layer capsules (e.g.\ the fixed
affinity $\rho^k_{ij}$ of \citet{smith-schut-gal-vanderwilk-21} parent
$i$ in layer $k$ for child $j$).  But if we consider a second-layer
capsule that is, for example, detecting the conjunction of the 3
strokes comprising a letter ``A'', then at different translations,
scales and rotations of the A it will need to connect to different
image level stroke features, so these connection affinities must
depend on transformation variables of the parent. This point
is illustrated in Fig.\ \ref{fig:supp_PCAE_equiv} of
sec.\ \ref{sec:learning} for the PCAE parts, and 
is also made in sec.\ 5.4 of \citet{smith-schut-gal-vanderwilk-21},
where it is shown that the parts decomposition used of a digit image
is not stable with respect to rotation of the digit.

\section{Learning Generative Capsule Models} \label{sec:learning}
Above we have assumed that the models are given, i.e.\ that
$F_{kn}$ and $\boldsymbol{m}_{kn}$ are known for each part $n$
and model $k$. One advantage of the interpretable nature of
the GCM is that one can use a ``curriculum learning'' approach
\citep{bengio-louradour-collobert-weston-09},
where individual models can first be learned, and then composed
together for inference in more complex situations.

For a given object, a key issue is the learning of the
decomposition into parts. For the constellations data this is not an
issue as the points (i.e.\ parts) are given, but for image data it
must be addressed. An important issue is the degree of flexibility of each
part---is it simply a fixed template, or are there internal degrees
of flexibility? An example of the former is 
non-negative  matrix factorization (NMF) \citep{lee-seung-99},
where, for example, aligned images of faces were decomposed into
non-negative basis functions (parts) that were combined with non-negative
coefficients. 
An example of a richer parts-based model is by 
\citet{ross-zemel-06},  who developed ``multiple cause factor analysis''
(MCFA) and applied it to faces, to learn the regions governed by each
part and the variability within each part.\footnote{In contrast to our
work they did not have a higher-level factor analyser to model
correlations between part appearances, but did allow variability in
the masks of the parts.} This work was also carried out on aligned,
vertically oriented face images, so it was not necessary to factor
out geometric transformations. For our work on faces, we made
use of the parts-based model from the ``PhotoFitMe'' work
described in section \ref{sec:parts_face}. This provided a
ready-made parts decomposition, but we learned a factor analysis model
on top to model the correlations between the parts.

\citet{kosiorek2019stacked} developed a Part Capsule Autoencoder
(PCAE) to learn parts from images, and applied it to MNIST images.
Each PCAE part is a template which can undergo an affine
transformation, and it has "special features" that were used to encode
the colour of the part.  If the overall model is to be equivariant to
geometric transformations, it is vital that the input part
decomposition is stable to such variation, otherwise the model is
building on shaky foundations.  However, we have observed that
the parts detected by PCAE are not equivariant to rotation.
Figure \ref{fig:supp_PCAE_equiv} shows that
the PCAE part decompositions inferred for a digit 4 are not stable to
different angles of rotation: notice e.g.\ in panel (a) how the part
coloured white maps to different parts of the 4 for
$45^{\circ}$-$180^{\circ}$ and $225^{\circ}$-$0^{\circ}$.
The details of this experiment are described in
Appendix \ref{sec:supp_PCAE}.

As identified above, a strength of the GCM is that
individual models can first be learned, before moving on to
more complex situations. We can start with an initial guess for the
object configuration, which can be chosen as one of the noisy
configurations from the training set. We then run variational
expectation maximization (variational EM; see sec.\ 6.2 in
\citealt*{jordan-ghahramani-jaakkola-saul-99}). In the E-step,
variational inference is run as in sec.\ \ref{sec:VI}
with this model to infer the $\bfy$ and
$Z$ variables on each training example. In the M-step, the
  variatonal distributions determined in the E-step are held fixed,
and the locations of the template points are updated so as to increase
the ELBO, summed over the training cases. Details of this update are
given in Appendix \ref{sec:app_learn_const} for the constellations
dataset, and experimental results are given in
sec.\ \ref{sec:constellation_learn_res}.

\section{Experiments and Results} \label{sec:expts_res}
We first provide experimental details and results for inference and
learning for the constellations data in
sec.\ \ref{sec:constellations}, and then give experimental
details and results for the faces data in
sec.\ \ref{sec:faces}.

\subsection{Constellations: experiments and results \label{sec:constellations}}
Below we provide details of the data generators, inference
methods and evaluation criteria for the constellations data in sec.\
\ref{sec:expts_constellation}, 
present results for inference on the constellations data in sec.\
\ref{sec:const_res}, and for learning constellation models in sec.\
\ref{sec:constellation_learn_res}.

\subsubsection{Constellations experiments \label{sec:expts_constellation}}
In order to allow fair comparisons, we use the same dataset generator
for geometric shapes employed by~\cite{kosiorek2019stacked}.  We
create a dataset of scenes, where each scene consists of a set of 2D
points, generated from different geometric shapes.  The possible
geometric shapes (templates) are a square and an isosceles triangle,
with parts being represented by the 2D coordinates of the vertices.
We use the same dimensions for the templates as used
by~\cite{kosiorek2019stacked}, side 2 for the square, and base and
height 2 for the triangle.  All templates are centered at $(0,0)$.  In
every scene there are at most two squares and one triangle, so
$N=11$. Each shape
is transformed with a random transformation to create a scene of 2D
points given by the object parts.  To match the evaluation of
\cite{kosiorek2019stacked}, all scenes are normalized so as the points
lie in $[-1,1]$ on both dimensions.  When creating the scene, we
select randomly (with probability 0.5) whether an object is going to
be present or not, but delete empty scenes.  A test set used for
evaluation is comprised of 450-460 non-empty scenes, based on 512
draws.

Additionally, we study how the methods compare when objects are
created from noisy templates.  We consider that the original templates
used for the creation of the images are corrupted with Gaussian noise
with standard deviation $\sigma$.  Once the templates are corrupted
with noise, a random transformation $\y_k$ is applied to obtain the
object $X_k$ of the scene.  As with the noise-free data, the objects
are normalized to lie in [-1,1] on both dimensions.

The CCAE is trained by creating random batches of 128 scenes as described
above and optimizing the objective function in~\eqref{eq:CCAE_loss}.
The authors run CCAE for $300K$ epochs, and when the parameters of the
neural networks are trained, they use their model on the test dataset
to generate an estimation of which points belong to which capsule, and
where the estimated points are located in each scene.

The variational inference approach allows us to model scenes where the
points are corrupted with some noise.  The annealing parameter $\beta$
controls the level of noise allowed in the model.  We use an annealing
strategy to fit $\beta$, increasing it every time the ELBO has
converged, up to a maximum value of $\beta_{max}=1$. We set the
hyperparameters of the model to $\bmu^g_0=\mathbf{0}$, $D^g_0 = I_4$,
$\lambda = 10^{4}$ and $a_{mnk} = \frac{1}{N}$.  We run
Algorithm~\ref{alg:variational_inference} with 5 different random
initializations of $R$ and select the solution with the best ELBO.
Similarly to~\citet{kosiorek2019stacked}, we incorporate a sparsity
constraint in our model, that forces every object to explain at least
two parts.  Once our algorithm has converged, for a given $k$ if any
$r_{mnk}>0.9$ and $\sum_m\sum_n r_{mnk} < 2$ it means that the model
has converged to a solution where object $k$ is assigned to less than
2 parts.  In these cases, we re-run
Algorithm~\ref{alg:variational_inference} with a new initialization of
$R$.  Notice that this is also related to the minimum basis size
necessary in the RANSAC approach for the types of transformations that
we are considering.

The implementation of SubsetMatch in
Algorithm~\ref{alg:geometric_hashing} considers all matches between
the predicted and the scene points where the distance between them is
less than $0.1$. Among them, it selects the matching with minimum
distance between scene and predicted points.

For both the variational inference algorithm and the RANSAC algorithm,
a training dataset is not necessary if we have prior knowledge of the
target shapes. This contrasts with SCAE, which learns from whole scenes.
If prior knowledge is not available, these shapes can be learned as
described in sec.\ \ref{sec:learning}, and illustrated in sec.\
\ref{sec:constellation_learn_res}.

Unfortunately we do not have access to the code employed
by~\citet{hinton2018matrix}, so we have been unable to make
comparisons with it.

{\bf Evaluation:} Three metrics are used to evaluate the performance
of the different methods: variation of information
\citep{meilua2003comparing}, adjusted Rand index
\citep{hubert1985comparing} and segmentation accuracy
\citep{kosiorek2019stacked}.  They are based on partitions of the
datapoints into those associated with each object, and those that are
missing.  Compared to standard clustering evaluation metrics, some
modifications are needed to handle the missing objects. Details are
provided in Appendix \ref{sec:supp_eval}. We also use an average scene
accuracy metric, where a scene is correct if the method returns the
full original scene, and is incorrect otherwise.

\begin{table}[h!]
\centering
\caption{Comparison between the different methods. For Segmentation
  Accuracy, Adjusted Rand Index and Scene Accuracy the higher the
  better. For Variation of Information the lower the better. Different
  levels of Gaussian noise with standard deviation $\sigma$ are
  considered.}

\vspace*{2mm}

\begin{tabular}{|c|c|c|c|c|}
\hline
Metric & Model & $\sigma$=0 & $\sigma$=0.1 & $\sigma$=0.25 \\
\hline
& CCAE & 0.828 & 0.754 & 0.623\\
{Segmentation} & GCM-GMM & 0.753 &     0.757 & 0.744\\
{Accuracy $\uparrow$} & GCM-DS & 0.899 &      0.882 & 0.785\\
& RANSAC & 1 & 0.992 & 0.965\\\hline
& CCAE & 0.599 & 0.484 & 0.248\\
{Adjusted} & GCM-GMM & 0.586 &     0.572 & 0.447\\
{Rand Index$\uparrow$}& GCM-DS & 0.740 &      0.699 & 0.498\\
& RANSAC & 1 & 0.979 & 0.914\\\hline
& CCAE & 0.481 & 0.689 & 0.988\\
{Variation of} & GCM-GMM & 0.478 &     0.502  & 0.677\\
{Information $\downarrow$}& GCM-DS &      0.299 & 0.359 & 0.659\\
& RANSAC & 0 & 0.034  & 0.135\\\hline
& CCAE & 0.365 & 0.138 & 0.033\\
Scene & GCM-GMM & 0.179 &       0.173  & 0.132\\
{Accuracy $\uparrow$} & GCM-DS &     0.664 & 0.603  & 0.377\\
& RANSAC & 1 & 0.961 & 0.843\\\hline
\end{tabular}
\label{t:main_results}
\end{table}

\subsubsection{Constellation Inference Experiments \label{sec:const_res}}
In Table~\ref{t:main_results} we show a comparison between CCAE, the
variational inference method with a Gaussian mixture
prior~\eqref{eq:p_Z} (GCM-GMM), with a DS prior over permutation
matrices~\eqref{eq:p_Z_true} (GCM-DS), and the RANSAC
approach for 
scenes without noise and with noise levels of $\sigma = 0.1, \; 0.25$
.\footnote{Code at: https://github.com/anazabal/GenerativeCapsules}
For GCM-GMM and GCM-DS we show the results where the
initial  $\beta = 0.05$. The effect of
different $\beta$ initializations is discussed below.

\begin{figure*}[t]
  \centering
  \includegraphics[width=0.9\linewidth]{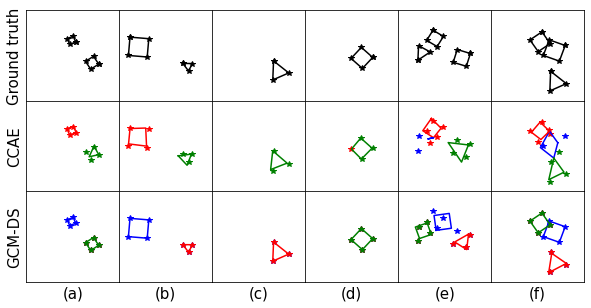}
  \caption{Reconstruction examples from CCAE and GCM-DS for noise-free
    data. The upper figures show the ground truth of the test images. The middle
    figures show the reconstruction and the capsule assignments (by
    colours) of CCAE. The lower figures show the
    reconstruction and the parts assignment of GCM-DS. Datapoints
      shown with a given colour are predicted
    to belong to the reconstructed object with the corresponding colour.}
  \label{fig:examples}
\end{figure*}

We see that GCM-DS improves over CCAE and GCM-GMM in all of the
metrics, with GCM-GMM being comparable to CCAE.  Interestingly, for
the noise-free scenarios, the RANSAC method achieves a perfect score
for all of the metrics.  Since there is no noise on the observations
and the method searches over all possible solutions of $\y_k$, it can
find the correct solution for any configuration of geometric shapes in
a scene.  For the noisy scenarios, all the methods degrade as $\sigma$
increases.  However, the relative performance between them remains the
same, with RANSAC performing the best, followed by GCM-DS and then
GCM-GMM.

Figure~\ref{fig:examples} shows some reconstruction examples from CCAE
and GCM-DS for the noise-free scenario.  In columns (a) and (b) we can
see that CCAE recovers the correct parts assignments but the object
reconstruction is inaccurate.  In (a) one of the squares is
reconstructed as a triangle, while in (b) the assignment between the
reconstruction and the ground truth is not exact.  For GCM-DS, if the
parts are assigned to the ground truth properly, {and there is no
  noise}, then the reconstruction of the object is perfect.  In column
(c) all methods work well.  In column (d), CCAE fits the square
correctly (green), but adds an additional red point.  In this case
GCM-DS actually overlays two squares on top of each other.  Both
methods fail badly on column (e).  Note that CCAE is not guaranteed to
reconstruct an existing object correctly (square or triangle in this
case).  In column (f) we can see that CCAE fits an irregular
quadrilateral (blue) to the assigned points, while GCM-DS obtains the
correct fit.
Additional examples for noisy cases with $\sigma=0.25$ are shown in
Appendix \ref{sec:noisy_examples}.

To assess the effect of in the initial value of $\beta$, we considered
6 values: 0.005, 0.01, 0.05, 0.1, 0.2, and 0.5. We found
that GCM-DS is always better than CCAE and GCM-GMM. As the
initial $\beta$ is increased, GCM-DS performs better across all the metrics. We
found that the performance of GCM-DS and GCM-GMM degrades with
$\beta > 0.1$.

We conducted paired t-tests between CCAE and GCM-GMM, GCM-DS and
RANSAC on the three clustering metrics for $\sigma = 0$ and initial
$\beta = 0.05$. The differences between CCAE and GCM-DS
are statistically significant with p-values less than $10^{-7}$, and
between CCAE and RANSAC with p-values less than $10^{-28}$.
For CCAE and GCM-GMM the differences are not statistically
significant.

\subsubsection{Learning  Constellations} \label{sec:constellation_learn_res}
Making use of the  interpretable nature of the GCM, we consider
learning each template (triangle or square) individually from noisy data, with
noise values of $\sigma = 0, \, 0.1, \, 0.25$. (Note that the full dataset
as generated contains 1/7 = 14\% single triangles, and 28\% single squares
which can  be selected simply based on the number of points.)

To compare the learned template to the ground-truth (GT), we have to bear
in mind that the learned template could be a rotated version, as this
will still be centered and have the correct scale. To remove this
rotational degree of freedom, we compute the Procrustes rotation
(see, e.g., \citealt*[sec.\ 14.7]{mardia-kent-bibby-79}) that best
aligns them. After this, we can compute the standardized mean
squared error (SMSE) $\frac{1}{N} \sum_{n=1}^N (\p_n - \p^{GT}_n)^T (\p_n - \p^{GT}_n)$.

We use $S = 64$ examples to train each template, and use one random
sample to seed the initial template (after translation, scaling and
rotation).  (In fact, for noise-free constellation data ($\sigma=0$),
a single observed triangle or square configuration serves as a
perfect template.) As $N$ is 3 or 4 in this case, we can use exact
inference over permutations, rather than the variational
approach. We initialized $\beta = 0.01$ and increased it after
  each iteration by a factor of two, similarly to the annealing
  strategy utilized in Algorithm
  \ref{alg:variational_inference}. The stopping criterion for the
  learning process was that the SMSE between two consecutive updates
  of the learned template should not be greater than $10^{-4}$.

\begin{figure}
    \centering
        \begin{tabular}{cc}
            \includegraphics[scale=0.4]{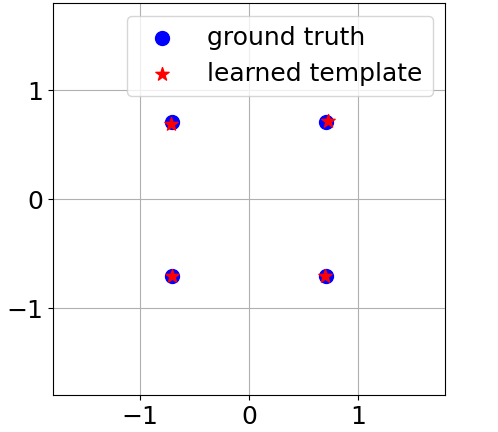} & \includegraphics[scale=0.4]{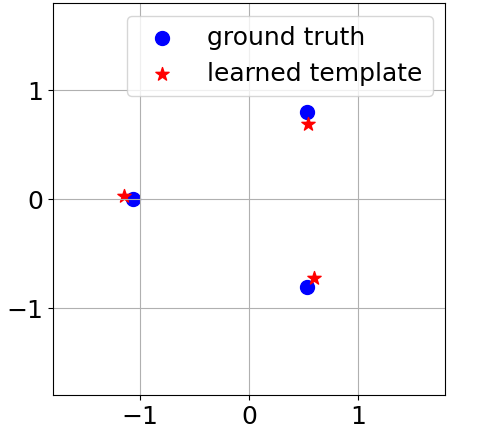}\\
            square, $\sigma=0.1$ & triangle, $\sigma=0.25$
        \end{tabular}
    \caption{Visualization of GT templates and Procrustes transformed learned templates, after training with scenes of different noise levels.}
    \label{fig:learned_templates}
\end{figure}

For the triangle object, the SMSE of the learned template was 
$4.1 \times 10^{-5}$ at $\sigma=0.1$, and $9.6 \times 10^{-3}$ at
$\sigma=0.25$.  For the square object, the corresponding SMSEs were
$1.8 \times 10^{-4}$ and $2.1 \times 10^{-2}$ respectively.
Figure \ref{fig:learned_templates} shows example learned templates
(after Procrustes rotation) compared to the GT templates. We see that
with only $S = 64$ examples, an accurate template can be
learned for each object. As expected, the error increases with
increasing $\sigma$. Note that $\sigma=0.25$ gives quite noticeably
distorted examples, see e.g.\ the examples in
Fig.\ \ref{fig:examples_sigma025} (top row).
In contrast, for the CCAE, the encoder network in the autoencoder is
not able to benefit from curriculum learning, and must tackle the full
problem for the start; \citet{kosiorek2019stacked}
used 300k batches of 128 scenes to train this model.

\subsection{Faces: experiments and results} \label{sec:faces}
Below we provide details of the data generator and inference
methods for the faces data in sec.\ \ref{sec:parts_face}, and
give inference results in sec.\ \ref{sec:face_res}.

\subsubsection{Parts-based face model} \label{sec:parts_face}
We have developed a novel hierarchical parts-based model for face
appearances. It is based on five parts, namely eyes, nose, mouth, hair
and forehead, and jaw (see Fig. \ref{fig:constandfaces}(b)). Each part
has a specified mask, and we have cropped the hair region to exclude
highly variable hairstyles. This decomposition is based on the
"PhotoFit Me" work and data of Prof.\ Graham Pike, see
\url{https://www.open.edu/openlearn/PhotoFitMe}.  For each part we
trained a probabilistic PCA (PPCA) model to reduce the dimensionality
of the raw pixels; the dimensionality is chosen so as to explain 95\%
of the variance. This resulted in dimensionalities of 24, 11, 12, 16
and 28 for the eyes, nose, mouth, jaw and hair parts respectively. We
then add a factor analysis (FA) model on top with latent variables
$\y^a_k$ to model the correlations of the PPCA coefficients across
parts. The dataset used (from PhotoFit Me) is balanced by gender
(female/male) and by race (Black/Asian/Caucasian), hence the
high-level factor analyser can model regularities across the parts,
e.g.\ wrt skin tone. $\x^a_{n}$ is predicted from $\y^a_k$ as
$F^a_{kn} \y^a_{k} + \boldsymbol{m}^a_{kn}$ as in (\ref{eq:pxa_supp}).
$\y^g_k$ would have an effect on the part appearance, e.g.\ by scaling
and rotation, but this can be removed by running the PPCA part
detectors on copies of the input image that have been rescaled and
rotated.

The ``PhotoFit Me'' project utilizes 7 different part-images for each
gender/race group, for each of the five part types. As a result, we
generated $7^5$  synthetic faces for each group, by combining these
face-parts, which led to a total of $100,842$ faces. All faces were
centered on a $224\times224$ pixel canvas. For each synthetic face we
created an appearance vector $\x^a_n$, which consisted of the stacked
vectors from the 5 different principal component subspaces. Finally,
we created a balanced subset from the generated faces ($18,000$
images), which we used to train a FA model. We tuned the latent dimension of
this model by training it multiple times with a different number of
factors, and finally chose 12 factors, where a knee in the
reconstruction loss on the face data was observed on a validation set.

To evaluate our inference algorithm we generated $224\times224$ pixel
scenes of faces. These consisted of 2, 3, 4 or 5
randomly selected faces from a balanced test-set of $7,614$ synthetic
faces, which were transformed with random similarity
transformations. The face-objects were randomly scaled down by a
minimum of 50\% and were also randomly translated and rotated, with
the constraint that all the faces fit the scene and did not overlap
each other. An example of such a scene is shown in Fig.\
\ref{fig:constandfaces}(c), and further examples are shown in 
Figure \ref{fig:face_inference_examples}. For each scene it is
  easy to determine the correct number of faces, as the number of faces
  present is equal to the count of each of the parts detected.
Afterwards, these two constraints were dropped to test the
ability of our model to perform inference with occluded 
parts, see Figure \ref{fig:face_inference_examples}(e) for
an example. These occluded scenes were comprised of 3 faces.
In our experiments we assume that the face parts
are detected accurately, i.e.\ as generated.

In the case of facial parts---and parts with both geometric and
appearance features in general---it only makes sense to assign the
observed parts $\x_m$ to template parts $\x_{kn}$ of the same type (e.g. an
observed ``nose'' part should be assigned only to a template ``nose''
part). We assume that this information is known, since the size of the
appearance vector of each part-type is unique. Thus it no
longer makes sense to initialize the assignment matrix uniformly for
all entries, but rather only for the entries that correspond to
templates of the observed part's type. Consequently,
\eqref{eq:log_rho} is only utilized for $m, n$ pairs of the same
type. Similarly to the constellation experiments, we initialized the
assignment matrix 5 times and selected the solution with the largest
ELBO.
  
In the experiments we evaluated the part assignment accuracy of the
algorithms. In a given scene, the assignment is considered correct if
all the observed parts have been correctly assigned to their
corresponding template parts with the highest probability. In
order to evaluate the prediction of the appearance features, we
measured the root mean square error (RMSE) between the input and
generated scenes.

\subsubsection{Face Experiments} \label{sec:face_res}

\begin{figure*}
  \centering
\begingroup
\setlength{\tabcolsep}{1pt} 
\renewcommand{\arraystretch}{1} 
\begin{tabular}{c ccc}
    Ground Truth & VI & RANSAC \\
    (a) \includegraphics[scale=0.28]{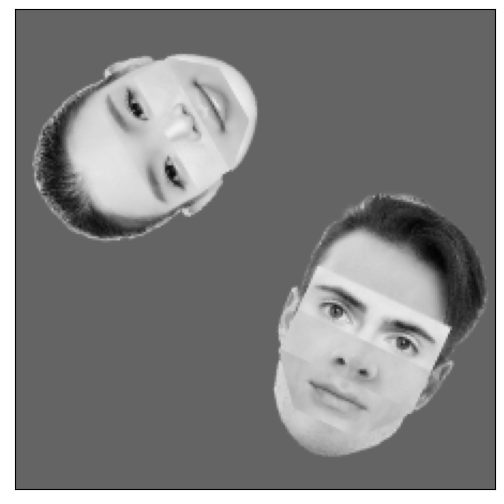} &
    \includegraphics[scale=0.28]{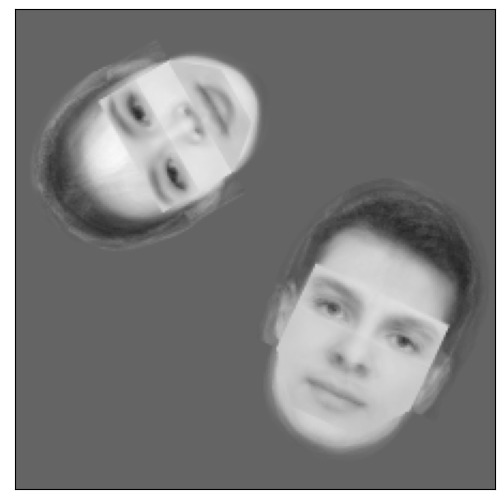} &
    \includegraphics[scale=0.28]{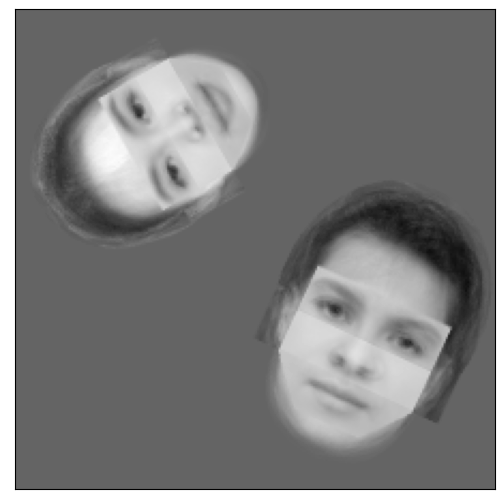} \\

    (b) \includegraphics[scale=0.28]{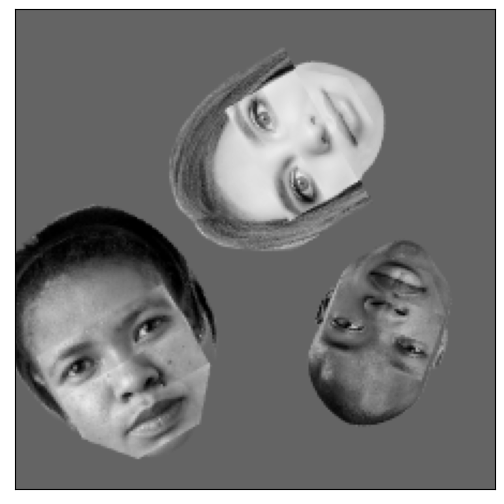} &
    \includegraphics[scale=0.28]{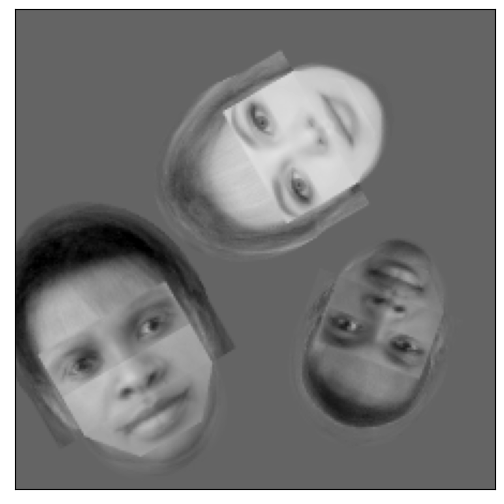} &
    \includegraphics[scale=0.28]{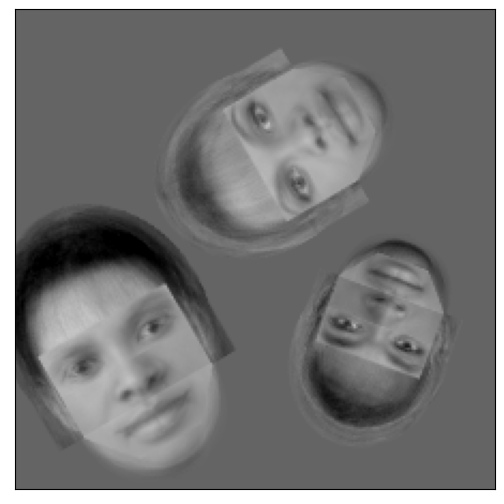} \\

    (c) \includegraphics[scale=0.28]{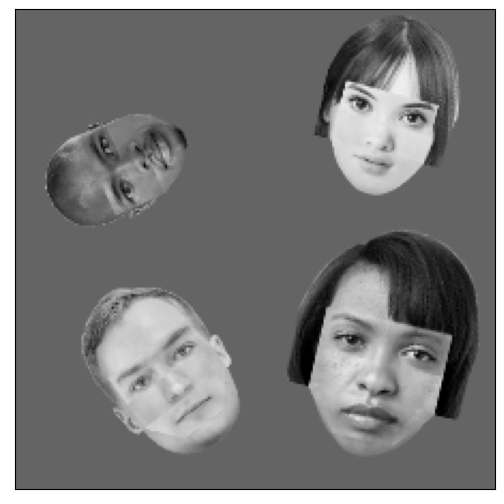} &
    \includegraphics[scale=0.28]{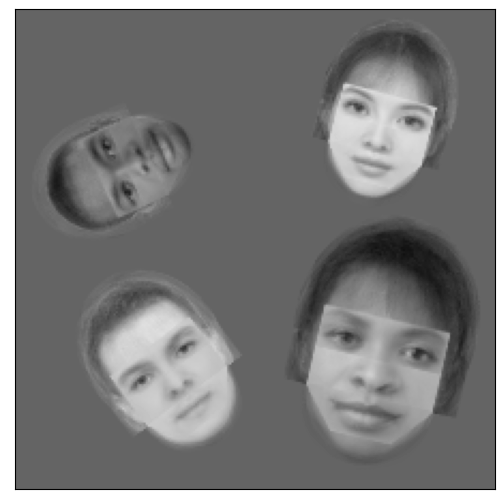} &
    \includegraphics[scale=0.28]{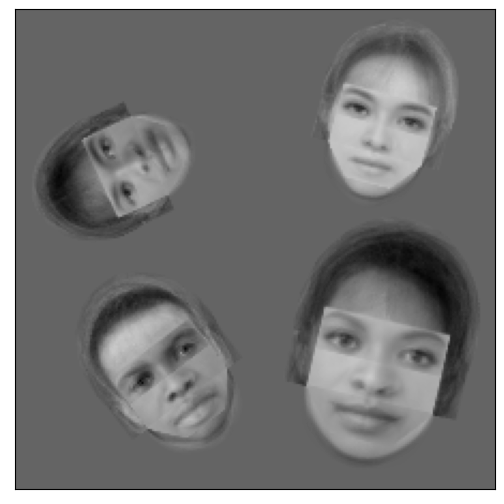} \\
 
    (d) \includegraphics[scale=0.28]{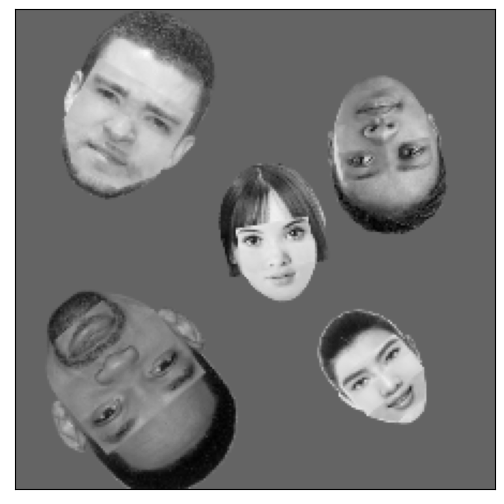} &
    \includegraphics[scale=0.28]{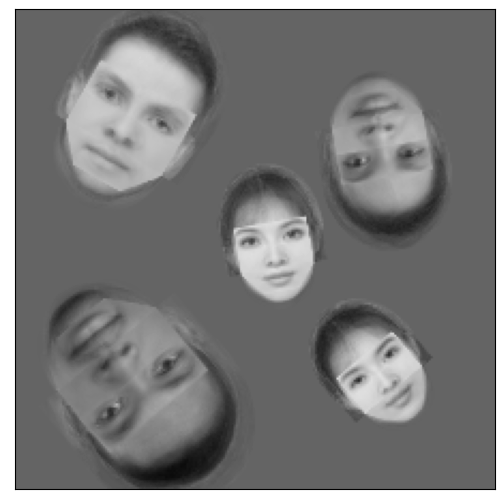} &
    \includegraphics[scale=0.28]{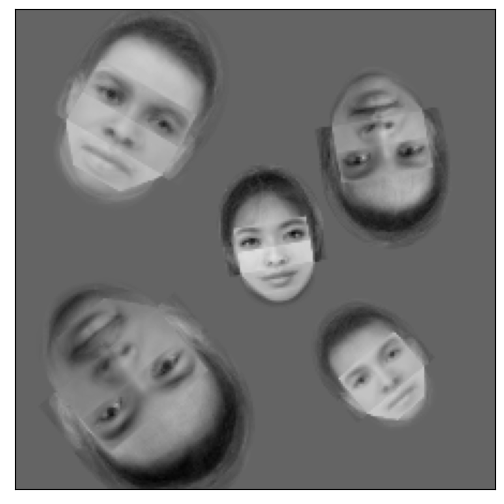}
    \\

    (e) \includegraphics[scale=0.28]{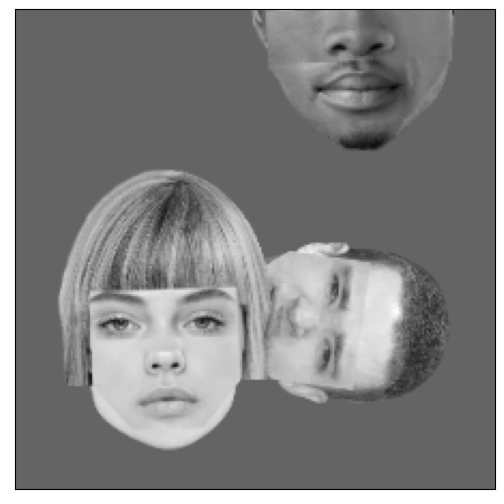} &
    \includegraphics[scale=0.28]{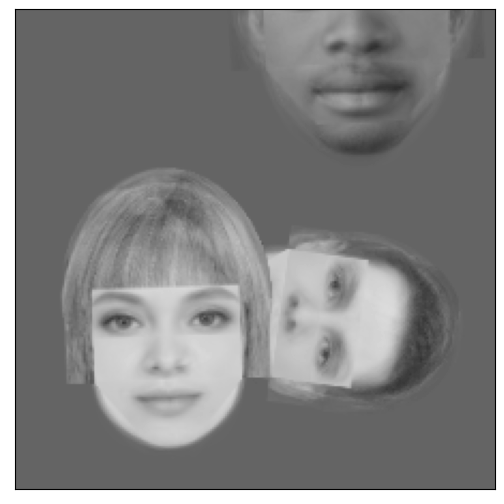} &
    \includegraphics[scale=0.28]{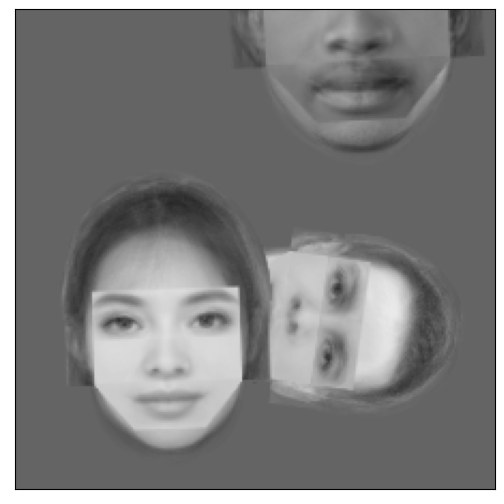} \\
      \end{tabular}
\endgroup
  \caption{Reconstruction examples with our Variational Inference (VI)
    algorithm and the RANSAC-type algorithm: (a) scene with 2 faces,
    (b) scene with 3 faces, (c) scene with 4 faces (d) scene with 5
    faces and (e) 3 faces with partially occluded faces. All faces
    have been randomly selected and transformed.}
\label{fig:face_inference_examples}
\end{figure*}

Firstly, the VI algorithm was evaluated on scenes of multiple,
randomly selected and transformed faces.\footnote{Code at: https://github.com/tsagkas/capsules}
For scenes with 2, 3, 4 and 5
faces, the assignment accuracy was 100\%, 100\%, 99.2\% and 93.7\%
respectively (based on 250 scenes per experiment). 
RANSAC gave 100\% accurate assignments in all four
cases. This is to be expected, since from each part the pose of the
whole can be predicted accurately. However, RANSAC's ability to infer the
appearance of the faces proved to be limited. More specifically, in
  250 instances uniformly distributed across scenes of 2, 3, 4 and 5
  faces, the VI algorithm had RMSE of $0.036\pm0.004$, while
  RANSAC scored $0.052\pm0.006$, with consistently higher error on \emph{all}
  scenes. This is illustrated in the examples of Figure
  \ref{fig:face_inference_examples}, where it is clear that RANSAC is
  less accurate in capturing key facial characteristics.
If inference for $\y^a_k$ is run as a post-processing step for
  RANSAC using all detected parts in an object, this difference disappears.
 
The supplementary material contains a movie showing the fitting of the
models to the data. It is not possible for us to make a fair
comparison with the SCAE algorithm on the faces data, as the PCAE
model used is not rich enough to model PCA subspaces.

Secondly, we evaluated the ability of our algorithm to perform
inference in scenes where some parts have been occluded, either by
overlapping with other faces or by extending outside of the scene. In 250
scenes with 3 partially occluded faces, both the VI and RANSAC
algorithms were fully successful in assigning the observed parts to
the corresponding template accurately; see Figure
\ref{fig:face_inference_examples}(e) for an example.

\section{Discussion} \label{sec:discuss}
In our experiments RANSAC was shown to often be an effective
alternative to variational inference (VI). This is particularly the
case when the basis in RANSAC is highly informative about the object.
For the constellations experiments this was the case, even when the
datapoints were corrupted by noise.
However, as we saw in sec.\ \ref{sec:face_res} for the faces data,
an individual part was less informative about the appearance
than the geometry, and so led to worse reconstructions unless
a post-processing step using all of the detected parts was used.
Also, RANSAC's  sampling-based inference may be less amenable to
neural-network style parallel implementations than VI.

Above we have described a \emph{generative} capsules model.
The key features of this approach are:
\begin{itemize}
\item The model is \emph{interpretable}, and thus admits the use of
  prior knowledge, e.g.\ if we already know some things about an
  object.  The formulation is also \emph{composable} in that models
  for individual objects can be learned separately, then combined
  together at inference time.
\item The variational inference algorithm is obtained directly
  from a generative model for the observations $X$. In contrast
  other leading formulations set up an objective to produce clusters in
  $\bfy$-space. 
\item The interpretable structure of the GCM allows other inference
  methods to be used, as demonstrated by our use of RANSAC.
\item The GCM conforms to the view, as promoted in
  \cite{kosiorek2019stacked}, that the input is regarded as a
  \emph{set} of parts. This formulation ensures that if the parts can
  be detected equivariantly, then the inferences for the objects will
  also be equivariant. This was demonstrated in the constellations
  and faces experiments.
\end{itemize}

As noted above, for the GCM to be equivariant to large transformations
of the input, the parts need to to be detected equivariantly. Some
capsules papers have used the affNIST
dataset\footnote{\url{https://www.cs.toronto.edu/~tijmen/affNIST/}},
but this only used small rotations of up to $\pm 20^{\circ}$.
\citet[sec.\ 5.1]{hinton2018matrix} did investigate the use of very
different viewpoints on the smallNORB dataset; while their capsules
results in Table 2 did outperform a competitor CNN, it is noticeable
that there is still a performance gap between novel and familiar
viewpoints. We have demonstrated (see Fig.\ \ref{fig:supp_PCAE_equiv})
that the PCAE decomposition is not equivariant to large rotations, and
similar observations have been made by
\citet{smith-schut-gal-vanderwilk-21} for their model.  Thus we
believe that further work on the equivariant extraction of parts is
necessary in order to achieve equivariant object recognition.

\subsection*{Acknowledgements}
We thank the anonymous referees for their helpful comments.
This work was supported in part by The Alan Turing Institute under
EPSRC grant EP/N510129/1.
For the purpose of open access, the authors have applied a Creative
Commons Attribution (CC BY)
licence to any Author Accepted Manuscript version arising from this submission.

\appendix

\section{Details for Variational Inference}\label{sec:appVI}

The evidence lower bound (ELBO) 
$L(q) = \bE_q[\log p(X,Y,Z) - \log q(Y,Z)]$
for this model is decomposed in three terms:
\begin{align}\label{eq:ELBO}
L(q) &=\bE_q[\log{p(X|Y,Z)}] - KL(q(Y)||p(Y)) - KL(q(Z)||p(Z)),
\end{align}
where $KL(q||p)$ is the Kullback-Leibler divergence between
distributions $q$ and $p$. 
The first term indicates how well the generative model $p(X|Y,Z)$ fits
the observations under our variational model $q(Y,Z)$:
\begin{align}
\bE_q[\log{p(X|Y,Z)}]  =  -\sum_{m=1}^M\sum_{k=1}^K\sum_{n=1}^{N_k} r_{mnk} \Big[
    \frac{d_{kn}}{2} \log{2\pi} +\frac{1}{2}\log{|\beta^{-1} D_{kn}|}
    + \Big.  \nonumber\\
\frac{\beta}{2}  (\x_{m} - F_{kn}\bmu_{k}- \boldsymbol{m}_{kn})^T D_{kn}^{-1}(\x_{m}
 - F_{kn}\bmu_{k} - \boldsymbol{m}_{kn})  + 
      \Big. \frac{\beta}{2} \text{trace}(F_{kn}^TD_{kn}^{-1}F_{kn}\Lambda_{k}^{-1}) \Big] . \label{eq:E_q_X}
\end{align}

The Kullback-Leibler divergence between the two Gaussian distributions $q(Y)$ and $p(Y)$ in our model has the following expression:
\begin{multline}
  KL(q(Y)||p(Y)) = \\
\frac{1}{2} \sum_{k=1}^K \left( 
\text{trace}(D^{-1}_0 \Lambda^{-1}_k ) - d_k + 
(\bmu_k -\bmu_0)^T D_0^{-1}(\bmu_k - \bmu_0) + 
\log{|D_{0}|} + \log{|\Lambda_k|} \right),
\end{multline}
where $d_k$ is the dimensionality of $\y_k$.

The expression for $KL(q(Z)||p(Z))$ is given by
\begin{align}\label{eq:KL_Z}
KL(q(Z)||p(Z)) &= \sum_{m=1}^{N}\sum_{k=1}^K\sum_{n=1}^{N_k}
r_{mnk}\log{\frac{r_{mnk}}{a_{mnk}}} .
\end{align}

\section{Learning for the Constellations  Model}\label{sec:app_learn_const}
We specialize the ELBO given in Appendix \ref{sec:appVI} for one
constellation, taking $K=1$ (and hence dropping the index $k$).  As
there are no appearance features, we drop the $g$ superscript on
$F_n$ and $\bfy$. Also the mean $\boldsymbol{m}_{kn}$ is only needed for the
appearance features and thus can be omitted. $D_n$ is specialized to
$\lambda^{-1} I_2$, we set $\beta = 1$ in this appendix and allow
$\lambda$ to vary,
and we assume $p(\bfy) \sim N(\bfzero, \Lambda_0^{-1})$.
Hence by specializing eq.\ \ref{eq:E_q_X} we obtain 
\begin{multline}\label{eq:E_q_Xconst}
\bE_q[\log{p(X|Y,Z)}]  =  \\ -\sum_{m=1}^M \sum_{n=1}^{N} r_{mn} \Big[
    \log \frac{2\pi}{\lambda}
    + \frac{\lambda}{2} (\x_{m} - F_{n}\bmu)^T (\x_{m} - F_{n}\bmu)
   +  \frac{\lambda}{2} \text{trace}(F_{n}^T F_{n} \Lambda^{-1}) \Big] ,
\end{multline}
where $q(\bfy) \sim N(\bmu, \Lambda^{-1})$, with $\Lambda$ and
$\bmu$ specialized from eqs.\ \ref{eq:Lambda_k} and \ref{eq:mu_k} as
\begin{align}
\Lambda = \Lambda_0  + \lambda \sum_{m=1}^M \sum_{n=1}^N r_{mn} F_n^T
F_n, \qquad
\bmu = \lambda \; \Lambda^{-1} (\sum_{m=1}^M \sum_{n=1}^N r_{mn} F_n^T
\x_m).
\label{eq:Lambda_mu_const}
\end{align}

Our goal in learning is to adapt the template parameters $\{ F_n \}$
so as to increase the variational log likelihood (ELBO) $L(q)$.
In the M-step of variational EM, the  distributions $q(\bfy)$ and
  $q(Z)$ (parameterized by $\bmu$, $\Lambda$ and $R$) are held fixed,
and the ELBO is optimized wrt $\{ F_n \}$.
Note that the  terms $KL(q(Y)||p(Y))$  and $KL(q(Z)||p(Z))$ do not
depend explicitly on $\{ F_n \}$, and hence any derivative of these KL
terms wrt $F_n$ will be zero. Thus these terms can be omitted when
optimizing the ELBO wrt $\{ F_n \}$.

The trace term in eq.\ \ref{eq:E_q_Xconst} can be simplified using
$\sum_m r_{mn} =1$, to give
\begin{equation}
\sum_{m=1}^M \sum_{n=1}^N r_{mn} \text{trace}(F_{n}^T F_{n}
\Lambda^{-1}) = \text{trace}( (\sum_{n=1}^N F_{n}^T F_{n})
\Lambda^{-1}) .
\end{equation}

It turns out that it is more convenient to write $F_n \bmu$ in terms
of $\p_n = (p_{nx}, p_{ny})^T$ and the mean transformation parameters
$\bmu= (\hat{t}_x, \hat{t}_y, \hat{s}_c, \hat{s}_s)^T$, where
$\hat{s}_c$ is the posterior mean of $s \cos \theta$, and 
$\hat{s}_s$ is the same for $s \sin \theta$. Hence we have that
\begin{align}
  F_n \bmu & = \begin{pmatrix} \hat{\t}_x \\  \hat{t}_y \end{pmatrix} +
               \begin{pmatrix} \hat{s}_c & \hat{s}_s \\ - \hat{s}_s &
                 \hat{s}_c \end{pmatrix}
               \begin{pmatrix}
                 p_{nx} \\
                 p_{ny} 
               \end{pmatrix}
            \defeq \; \hat{\t} + \hat{T} \p_n .
\end{align}

Hence $\x_m - F_n \bmu = \x_m - \hat{\t} - \hat{T} \p_n =
\tilde{\x}_m - \hat{T} \p_n$, where $\tilde{\x}_m = \x_m - \hat{\t}$,
and the quadratic form $(\x_{m} - F_{n}\bmu)^T (\x_{m} -
F_{n}\bmu)$ can be rewritten as $(\tilde{\x}_m - \hat{T} \p_n)^T
(\tilde{\x}_m - \hat{T} \p_n)$.

We can now rewrite the term $Q = \sum_{m} \sum_n r_{mn} 
(\x_{m} - F_{n}\bmu)^T (\x_{m} - F_{n}\bmu)$ in
eq.\ \ref{eq:E_q_Xconst} as
\begin{align}
  Q &= \sum_{m} \sum_n r_{mn} (\tilde{\x}_m - \hat{T} \p_n)^T
  (\tilde{\x}_m - \hat{T} \p_n) \\
    &=  \sum_{m} \sum_n r_{mn} \left( \p_n^T \hat{T}^T \hat{T} \p_n -
  \p_n^T \hat{T}^T \tilde{\x}_m - \tilde{\x}_m^T \hat{T} p_n +
  \tilde{\x}_m^T \tilde{\x}_m \right) .
\end{align}
Using $\sum_m r_{mn} = 1$ and defining $\tilde{\x}_n^r = \sum_{m}
r_{mn} \tilde{\x}_m$, we obtain
\begin{equation}
  Q = \sum_n \p_n^T \hat{T}^T \hat{T} \p_n -
  \p_n^T \hat{T}^T \tilde{\x}^r_n - (\tilde{\x}^r_n)^T \hat{T} \p_n +
  (\sum_{m,n} r_{mn} \tilde{\x}_m^T \tilde{\x}_m) .
\end{equation}
This can be further simplified by noting that $\hat{T}^T \hat{T} =
\hat{s}^2 I_2$, where $\hat{s}^2 = \hat{s}^2_c + \hat{s}^2_s$.

The above derivation is all for one example $X$. Now summing $Q$ over all
training examples $\{ X_i \}$ we obtain
\begin{equation}
  Q^{tot} = \sum_{i} Q_i = \sum_{i} \sum_n \left( \hat{s}^2_i \p_n^T \p_n -
  \p_n^T \hat{T}^T \tilde{\x}^r_{ni} - (\tilde{\x}^r_{ni})^T \hat{T} \p_n +
  \sum_{m} r_{mn} \tilde{\x}_{mi}^T \tilde{\x}_{mi} \right), \label{eq:Qtot}
\end{equation}
where $\tilde{\x}^r_{ni}$ denotes $\tilde{\x}^r_{n}$ in the $i$th
example, and similarly for $\tilde{\x}_{mi}$.

Now consider the trace term $S_i =  \text{trace}( (\sum_{n=1}^N
F_{n}^T F_{n} )\Lambda_i^{-1})$, where $\Lambda_i$ is the precision
matrix for $\y$ on the $i$th example. We have that
\begin{align}
  F_n^T F_n =   \begin{pmatrix}
    1 & 0 \\
    0 & 1 \\
    p_{nx} & p_{ny} \\
    p_{ny} & - p_{nx}
    \end{pmatrix}
  \begin{pmatrix}
    1 & 0 & p_{nx} & p_{ny} \\
    0 & 1 & p_{ny} & - p_{nx}
  \end{pmatrix}
  =
  \begin{pmatrix}
    1 & 0 &  p_{nx} & p_{ny} \\
    0 & 1 & p_{ny} & - p_{nx} \\
    p_{nx} & p_{ny} & p^2_{nx} +p^2_{ny} & 0 \\
    p_{ny} & - p_{nx} & 0 & p^2_{nx} +p^2_{ny}
  \end{pmatrix}     .
\end{align}
Assume that the template is centered so that $\sum_n p_{nx} =
\sum_n p_{ny} = 0$, and scaled so that $\sum_n p^2_{nx} +p^2_{ny}  =N$.
Hence we have that $\sum_n F_n^T F_n = N I_4$.  From
eq.\ \ref{eq:Lambda_mu_const} and taking $\Lambda_0 = \alpha I_4$
we have
\begin{equation}
  \Lambda = \Lambda_0 + \sum_m \sum_n r_{mn} F_n^T F_n =
(\alpha+ \lambda N) I_4 .
\end{equation}
Hence
\begin{equation}
\text{trace}( (\sum_{n=1}^N F_{n}^T F_{n} )\Lambda^{-1}) = N
\text{trace} (\Lambda^{-1}) = \frac{4 N}{\alpha + \lambda N} .
\end{equation}
Crucially, this term is \emph{independent} of $\{ \p_n \}$, as long as the
template is centered and scaled correctly. Hence this term can be
ignored when optimizing the $\p_n$s.

Thus optimizing the ELBO wrt $\p_n$ comes down to optimizing $Q^{tot}$
wrt $\p_n$. Differentiating eq.\ \ref{eq:Qtot} wrt $\p_n$ and setting
it equal to zero we obtain
\begin{equation}
\frac{\partial {Q^{tot}}}{\partial {\p_n}} = \sum_i \left( 2
\hat{s}^2_i \p_n -  2 \hat{T}_i^T \tilde{\x}^r_{ni} \right) = 0,
\end{equation}
which gives the update formula
\begin{equation}
  \p_n = \frac{1}{\sum_i \hat{s}_i^2} \sum_i \hat{T}^T_i
  \tilde{\x}^r_{ni} .
\end{equation}
It can also be shown that $\hat{T}^T = \hat{s}^2 \hat{T}^{-1}$,
yielding the update equation
\begin{equation}
  \p_n = \frac{\sum_i \hat{s}_i^2 \hat{T}^{-1}_i \tilde{\x}^r_{ni}}{\sum_i \hat{s}_i^2} .
\end{equation}
This is quite intuitive---we first remove the effect of the
translation $\hat{\t}$ by computing $\tilde{\x}_m$, then take into
account the weighted assignments $r_{mn}$ to give
$\tilde{\x}^r_{ni}$, and then apply $\hat{T}^{-1}_i$ to remove the
effect of the scaling and rotation. The summations are weighted by
$\hat{s}_i^2$, which has the effect giving higher weight to examples
with larger scale, where the relative effect of the noise
$N(0,\lambda^{-1})$ is smaller.

One can also re-estimate $\lambda$ using the variational EM
algorithm. Differentiating eq.\ \ref{eq:E_q_Xconst} wrt $\lambda$ we
obtain
\begin{equation}
 \frac{1}{\lambda} = \frac{1}{2 N S} \sum_{i=1}^S \sum_{m,n}  r^i_{mn} (\x_{mi} -
 F_n \bmu_i)^T (\x_{mi} -  F_n \bmu_i) , \label{eq:reest_lambda}
\end{equation}
where $S$ denotes the number of examples.
As $\lambda^{-1}$ is a variance, this equation makes sense in terms of
an average of squared residuals. In the derivation of eq.\ \ref{eq:reest_lambda}
the dependence of the trace term $4 N \lambda/(\alpha + N \lambda)$
on $\lambda$ has been omitted, as for $N \lambda \gg \alpha$ this
dependence is negligible.

\section{Evaluation metrics for the constellations data} \label{sec:supp_eval}
In a given scene $X$ there are $M$ points, but we know that there are
$N \ge M$ possible points that can be produced from all of the
templates.  Assume that $K' \le K$ templates are active in this scene.
Then the points in the scene are labelled with indices $1, \ldots,
K'$, and we assign the missing points index $0$.  Denote the ground
truth partition as $V = \{ V_0, V_1, \ldots V_{K'} \}$.  An
alternative partition output by one of the algorithms is denoted by
$\hat{V} = \{ \hat{V}_0, \hat{V}_1, \ldots \hat{V}_{\hat{K}'} \}$.
The predicted partition $\hat{V}$ may instantiate objects or points
that were in fact missing, thus it is important to handle the missing
data properly.

In Information Theory, the \textbf{variation of information}
(VI)~\citep{meilua2003comparing} is a measure of the distance between
two partitions of elements (or clusterings).  For a given set of
elements, the variation of information between two partitions $V$ and
$\hat{V}$, where $N = \sum_i |V_i| = \sum_j |\hat{V}_j|$ is defined
as:
\begin{equation}
VI(V,\hat{V}) = -\sum_{i,j} r_{ij}\left[\log{\frac{r_{ij}}{p_i}} + \log{\frac{r_{ij}}{q_j}} \right]
\end{equation}
where $r_{ij} = \frac{|V_i \cap \hat{V}_j|}{N}$, $p_i =
\frac{|V_i|}{N}$ and $q_j = \frac{|\hat{V}_j|}{N}$.  In our
experiments we report the average variation of information of the
scenes in the dataset.

The \textbf{Rand index}~\citep{rand1971objective} is another measure
of similarity between two data clusterings.  This metric takes pairs
of elements and evaluates whether they do or do not belong to the same
subsets in the partitions $V$ and $\hat{V}$
\begin{equation}
RI = \frac{TP + TN}{TP +TN +FP +FN},
\end{equation}
where TP are the true positives, TN the true negatives, FP the false
positives and FN the false negatives.  The Rand index takes on values
between 0 and 1.  We use instead the \textbf{adjusted Rand index}
(ARI) \citep{hubert1985comparing}, the corrected-for-chance version of
the Rand index.  It uses the expected similarity of all pair-wise
comparisons between clusterings specified by a random model as a
baseline to correct for assignments produced by chance.  Unlike the
Rand index, the adjusted Rand index can return negative values if the
index is less than the expected value.  In our experiments, we compute
the average adjusted Rand index of the scenes in our dataset.

The \textbf{segmentation accuracy} (SA) is based on obtaining the
maximum bipartite matching between $V$ and $\hat{V}$, and was used by
\citet{kosiorek2019stacked} to evaluate the performance of CCAE.  For
each set $V_i$ in $V$ and set $\hat{V}_j$ in $\hat{V}$, there is an
edge $w_{ij}$ with the weight being the number of common elements in
both sets.  Let $W(V,\hat{V})$ be the overall weight of the maximum
matching between $V$ and $\hat{V}$.  Then we define the average
segmentation accuracy as:
\begin{equation}
SA = \sum_{i=1}^I  \frac{W(V_i,\hat{V}_i)}{W(V_i,V_i)} = \frac{1}{N}\sum_{i=1}^I W(V_i,\hat{V}_i),
\end{equation}
where $I$ is the number of scenes.
Notice that $W(V_i,V_i)$ represents a perfect assignment of the ground
truth, both the observed and missing subsets, and thus $W(V_i,V_i) =
N$.

There are some differences on how we compute the SA metric compared
to~\cite{kosiorek2019stacked}.  First, they do not consider the
missing points as part of their ground truth, but as we argued above
this is necessary.  They evaluate the segmentation accuracy in terms
of the observed points in the ground truth, disregarding possible
points that were missing in the ground truth but predicted as observed
in $\hat{V}$.  Second, they average the segmentation accuracy across
scenes as
\begin{equation}
  SA =   \frac{\sum_{i=1}^I W(V_i,\hat{V}_i)}{\sum_{i=1}^I W(V_i,V_i)} .
\label{eq:SAkos}  
\end{equation}
For them, $W(V_i,V_i) = M_i$, where $M_i$ is the number of points
present in a scene.  In our case, both averaging formulae are
equivalent since our $W(V_i,V_i)$ is the same across scenes.

\section{Failures of rotation equivariance for SCAE} \label{sec:supp_PCAE}
We trained the SCAE model with digit ``4'' images from the training
set of the MNIST dataset\footnote{http://yann.lecun.com/exdb/mnist .},
after they had been uniformly
rotated by up to $360^\circ$ and uniformly translated by up to 6
pixels on the x and y axes. Since we used a single class in the
dataset we modified the SCAE architecture to use only a single object
capsule.

We repeated the training of SCAE multiple times for 8K epochs, and
collected distinct sets of learned $11\times11$ parts that the digit
``4'' can be decomposed into. Two example part-sets are shown in
Fig.\ \ref{fig:supp_PCAE_equiv}. We then evaluated PCAE's
ability to detect these parts in MNIST digit ``4'' images that had
been rotated by multiples of $45^\circ$. 
Our results indicate that the PCAE model is not equivariant
to rotations. This is apparent from Fig. \ref{fig:supp_PCAE_equiv},
where the learned parts
are inconsistently assigned to the
regions of the digit-object, depending on the angle of rotation. We
hypothesize that this phenomenon stems from the fact that PCAE seems
to generate parts that are either characterized by an intrinsic
symmetry, and thus their pose is ambiguous, as in the line features of
part-set-a in Fig.\ \ref{fig:supp_PCAE_equiv});
or pairs of parts that
are transformed versions of themselves, and thus can be used
interchangeably (e.g. the first and third templates of part-set-b in
Fig.\ \ref{fig:supp_PCAE_equiv}). This leads to identifiability issues,
where the object can be decomposed into its parts in numerous ways.

\section{Examples of Noisy Cases}\label{sec:noisy_examples}

\begin{figure*}[t]
  \centering
  \includegraphics[width=0.85\linewidth]{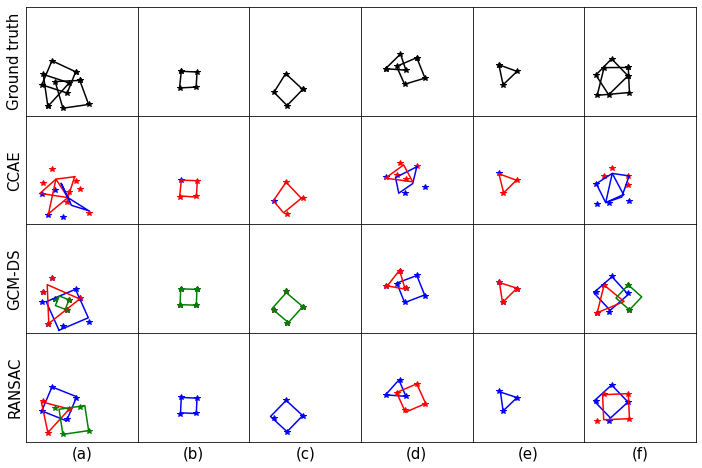}
  \caption{Reconstruction examples from CCAE, GCM-DS and RANSAC with Gaussian noise $\sigma=0.25$.}
  \label{fig:examples_sigma025}
\end{figure*}

Figure~\ref{fig:examples_sigma025} 
shows several examples of objects generated from noisy templates
with a corruption level of $\sigma=0.25$.
GCM-DS and RANSAC do not allow for deformable objects to try to fit
the points exactly, contrary to CCAE.  Both methods try to find the
closest reconstruction of the noisy points in the image by selecting
the geometrical shapes that are a best fit to those points.
Nonetheless, both methods can determine that which parts belong
together to form a given object, even when the matching is not
perfect.

\cut{
\begin{figure*}[t]
  \centering
  \caption{Reconstruction examples from CCAE, GCM-DS and RANSAC with Gaussian noise $\sigma=0.1$.}
  \includegraphics[width=0.7\linewidth]{Figures/example_images_01.png}
  \label{fig:examples_sigma01}
\end{figure*}
}

\vskip 0.2in
\bibliographystyle{apalike}

\end{document}